\documentclass[lettersize,journal]{IEEEtran}
\usepackage{amsmath,amsfonts}
\usepackage{algorithmic}
\usepackage{algorithm}
\usepackage{array}
\usepackage[caption=false,font=normalsize,labelfont=sf,textfont=sf]{subfig}
\usepackage{textcomp}
\usepackage{stfloats}
\usepackage{url}
\usepackage{verbatim}
\usepackage{graphicx}
\usepackage{cite}
\usepackage{xcolor}
\usepackage{soul,xcolor}

\def\BibTeX{{\rm B\kern-.05em{\sc i\kern-.025em b}\kern-.08em
    T\kern-.1667em\lower.7ex\hbox{E}\kern-.125emX}}
\begin{document}

\title{Simultaneous Estimation of Hand Configurations and Finger Joint Angles using Forearm Ultrasound}

\author{Keshav Bimbraw, Christopher J. Nycz, Matt Schueler, Ziming Zhang, and Haichong K. Zhang
\thanks{Keshav Bimbraw is currently working as an Augmented Human Sensing Intern under Data and Devices Group in the Artificial Intelligence Research Lab at Nokia Bell Labs, 600 Mountain Ave bldg 5, New Providence, NJ 07974, USA. He is also a Ph.D. student in Robotics Engineering at Worcester Polytechnic Institute, 100 Institute Rd, Worcester, MA 01609, USA. (e-mail: kbimbraw@wpi.edu).}
\thanks{Christopher J. Nycz is a research scientist working at PracticePoint, Worcester Polytechnic Institute, 50 Prescott Street, Worcester, MA 01605, USA. (e-mail: cjnycz@wpi.edu).}
\thanks{Matthew Schueler works as a Firmware Engineer at Amazon Robotics, 50 Otis St, Westborough, MA 01581, USA. (e-mail: mschueler@wpi.edu).}
\thanks{Ziming Zhang is with the Vision, Intelligence, and System Laboratory (VISLab) at Worcester Polytechnic Institute, 100 Institute Rd, Worcester, MA 01609, USA. (e-mail: zzhang15@wpi.edu).}
\thanks{Haichong K. Zhang is with the Medical FUSION Lab at Worcester Polytechnic Institute, 100 Institute Rd, Worcester, MA 01609, USA. (e-mail: hzhang10@wpi.edu).}}
\markboth{IEEE TRANSACTIONS ON MEDICAL ROBOTICS AND BIONICS,~Vol.~xx, No.~x, August~xxxx}%
{Bimbraw \MakeLowercase{\textit{et al.}}: Simultaneous Estimation of Hand Configurations and Finger Joint Angles using Forearm Ultrasound}


\maketitle

\begin{abstract}
With the advancement in computing and robotics, it is necessary to develop fluent and intuitive methods for interacting with digital systems, augmented/virtual reality (AR/VR) interfaces, and physical robotic systems. Hand motion recognition is widely used to enable these interactions. Hand configuration classification and metacarpophalangeal (MCP) joint angle detection is important for a comprehensive reconstruction of hand motion. Surface electromyography (sEMG) and other technologies have been used for the detection of hand motions. Forearm ultrasound images provide a musculoskeletal visualization that can be used to understand hand motion. Recent work has shown that these ultrasound images can be classified using machine learning to estimate discrete hand configurations. Estimating both hand configuration and MCP joint angles based on forearm ultrasound has not been addressed in the literature. In this paper, we propose a convolutional neural network (CNN) based deep learning pipeline for predicting the MCP joint angles. The results for the hand configuration classification were compared by using different machine learning algorithms. Support vector classifier with different kernels, multi-layer perceptron, and the proposed CNN have been used to classify the ultrasound images into 11 hand configurations based on activities of daily living. Forearm ultrasound images were acquired from 6 subjects instructed to move their hands according to predefined hand configurations. Motion capture data was acquired to get the finger angles corresponding to the hand movements at different speeds (0.5 Hz, 1 Hz, \& 2 Hz) for the index, middle, ring, and pinky fingers. Average classification accuracy of 82.7 $\pm$ 9.7\% for the proposed CNN and over 80\% for SVC for different kernels was observed on a subset of the dataset. An average RMSE of 7.35$^{\circ}\pm$1.3$^{\circ}$ was obtained between the predicted and the true MCP joint angles. A low latency (6.25 - 9.1 Hz) pipeline has been proposed for estimating both MCP joint angles and hand configuration aimed at real-time control of human-machine interfaces.
\end{abstract}


\section{Introduction}
\IEEEPARstart{S}{mart} and intuitive upper limb interaction with physical and non-physical worlds (AR/VR) has an emerging interest in the human-computer interaction research community and has numerous interfacing and control applications \cite{tarasenko2020artificial, vuletic2019systematic, jung2021skin, zhu2020haptic, li2021survey,  makhataeva2020augmented}. Upper limb motor dexterity in humans is possible with a highly advanced neuromuscular machinery developed over millions of years of evolution from primitive primates with prehensile appendages to the upper limbs. Human upper limbs have the maximum proportion of motor-sensory innervation in the human body and are used to interact with digital systems, augmented/virtual reality (AR/VR) interfaces, and physical robotic systems\cite{brown1994human}. While carrying out these activities of daily living (ADLs), the primary joint for the fingers is the metacarpophalangeal (MCP) joint and it plays a major role in manipulation tasks \cite{vuletic2019systematic}. It is thus important to estimate hand configurations and MCP joint angles to facilitate this interaction.

\begin{figure*}[ht]
    \centering
    \includegraphics[scale=0.53]{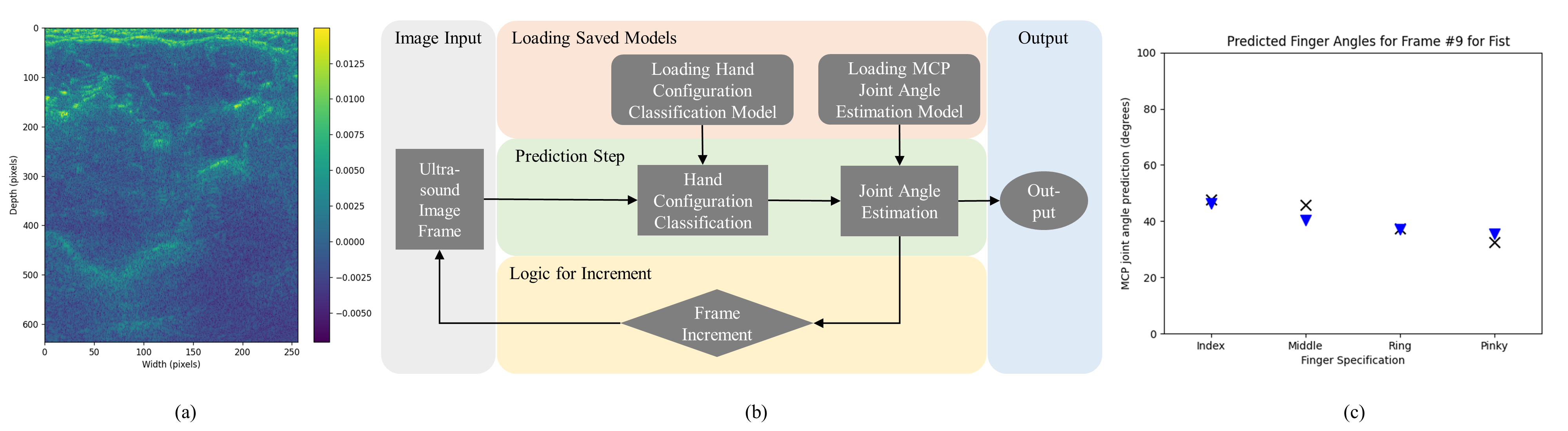}
    \caption{The combined pipeline for hand configuration classification and MCP joint angle prediction: (a) An example of the input ultrasound image, (b) the data processing pipeline, and (c) an example of the output prediction plot. The crosses are the true values and the inverted triangles are the predicted values.}
    \label{fig:fig_1}
\end{figure*}

Various technologies have been used for hand configuration estimation and interpreting hand movements through sensing regions on the hand or the forearm \cite{cheok2019review}. These include surface electromyography (sEMG) \cite{ahsan2009emg}, force myography (FMG) \cite{jiang2017exploration}, vision-based approaches \cite{oudah2020hand}, resistive hand gloves \cite{guo2021human}, depth information-based approaches \cite{suarez2012hand} and WiFi sensing \cite{ahmed2020device}. However, these state-of-the-art methods have limitations. Placing sensors on the fingers, such as bend sensors and motion sensors can limit mobility and usage of the hands to their full extent for the individuals wearing such sensors \cite{dipietro2008survey}. Vision-based methods for hand state and orientation detection are often sensitive to poor resolution, frame rate, drastic illumination conditions, changing weather conditions and occlusions \cite{rautaray2015vision, chakraborty2018review}. 

Biological signals from the forearm can be used as an alternative to understanding hand movement without impeding the user and preventing difficult and awkward hand movements \cite{esposito2021biosignal}. sEMG has been used for the recognition of hand motion and can facilitate human-robot-environment integration with the help of an intelligent robot perception system  \cite{li2020review}. It has been researched for its use as a control input to prosthetics and powered orthotics \cite{graupe1982multifunctional}. Various combinations of sensors with sEMG have also been explored for upper and lower limb motion prediction \cite{lee2020image, kahanowich2021robust, dong2021soft, rabe2020use}. Araki et al. used sEMG to measure the middle finger metacarpophalangeal and proximal inter-phalangeal joint angles and used it to control a robotic finger \cite{araki2012artificial}. Shrirao et al. used sEMG to predict index finger metacarpophalangeal joint angle while performing flexion-extension rotation of the index finger \cite{shrirao2009neural}. Wang et al. used sEMG to predict finger joint angles during different grasps \cite{wang2020semg}. Deep Learning has also been used for facilitating sEMG-based human computer interaction \cite{xiong2021deep}. Tosin et al. provide a current review of sEMG-based upper limb movement classifiers \cite{tosin2022semg} and Palumbo et al. provide a review of biopotential signal monitoring in rehabilitation \cite{palumbo2021biopotential}. Several commercial solutions have been created based on sEMG data processing for hand motion prediction in the context of prosthetics including Coapt Gen2 and Myo Plus Ottobock. However, these commercial systems have not been widely accepted to replace the traditional direct control \cite{zhou2022non}.
sEMG is sensitive to muscle fatigue because of long-term muscle movements and works best for fast movements of the hand which can inhibit human-like control of digital and virtual interfaces, and prosthetics/orthotics \cite{li2020review, shrirao2009neural}. Muscle fatigue produces a decrease in the mean frequency and has a variable effect on the sEMG signal amplitude \cite{ortega2018fatigue}. These biological signal limitations in addition to signal contamination due to motion artifact, and electromagnetic, and environmental interference might introduce variation in time, frequency, and statistical properties of the signal and are an impediment to easy adoption \cite{tosin2022semg}.
\subsection{Related Works}
While sEMG can be used to give an estimate of the muscle activation for hand and finger movements, ultrasound imaging of the forearm can give a 2-dimensional visualization of the musculoskeletal structure of the forearm. Ultrasound imaging of the forearm, or Sonomyography, has been explored as an alternative sensing modality that has been shown to be capable of identifying different hand gestures and finger movements from ultrasound data with a combination of image processing and classification algorithms \cite{bimbraw2020towards, zheng2006sonomyography, chen2010sonomyography, shi2010feasibility}. Sonomyography has also been used to classify several distinct hand configurations and controlling robotic mechanisms \cite{bimbraw2020towards}. Akhlagi et al. demonstrated the classification of 15 hand motions with an average classification accuracy of 91\% using Nearest Neighbor classification algorithms \cite{akhlaghi2015real}. 

Yang et al. described the estimation of simultaneous and distinct wrist pronation-supination and hand opening-closing based on ultrasound imaging using subclass discriminant analysis (SDA) and principal component analysis \cite{yang2020wearable}. They were able to achieve a wrist rotation classification accuracy of 99.2\% and a hand gesture classification of 92.8\%. McIntosh et al. classified 10 hand gestures with an average accuracy percentage of over 99\% based on small ultrasound data sizes using support vector machines (SVMs) and multi-layer perceptrons (MLPs) \cite{mcintosh2017echoflex}. Huang et al. compared the classification performance based on ultrasound and sEMG data for 14 finger motions \cite{huang2017ultrasound}. They treated their ultrasound data as A-mode ultrasound signals and did not consider the spatial features which could be acquired from the full B-mode ultrasound images since according to them it was redundant.
\subsection{Contributions}
While discrete hand configuration classification using forearm ultrasound images has been implemented previously, there remains to be a need for the prediction of the finger joint angles continuously for effectively controlling systems and environments. This can be facilitated by using deep learning based algorithms and techniques to estimate hand configurations relevant to ADLs, and MCP joint angles while attaining them. Since hand gestures are a primary way to convey information because of a high degree of differentiation, flexibility, and efficiency of information transmission for human-machine interfacing, it is important to estimate them for effective human-machine interaction \cite{guo2021human}. This is especially useful for developing technology as a communication medium for people who cannot communicate verbally \cite{cheok2019review}. There is also an active research community focused on robotics applications related to the development of rehabilitation tools and robots which can assist with upper limb rehabilitation \cite{gerlich2007gesture, gu2022review}. This work can be beneficial to these domains because of a direct estimation from a physiological visualization of the musculoskeletal structure of the forearm. This paper discusses a pipeline for estimating these hand configurations and MCP joint angles. A convolutional neural network (CNN) has been proposed for estimating MCP joint angles. The hand classification accuracy results have been compared by using different machine learning algorithms. Results for hand configuration classification using a support vector classifier (SVC) with different kernels, a multi-layer perceptron (MLP), and the proposed CNN have been discussed. Results from simultaneous hand configuration and angle prediction pipeline have also been described.

\section{Methods}
A low-latency solution to demonstrate the performance of the combined hand configuration classification and finger angle prediction pipeline is presented. SVC, MLP, and the proposed CNN were used for hand configuration classification, and the same CNN architecture was used for continuously predicting MCP joint angles based on the forearm ultrasound.
\subsection{Combined System}
The combined system demonstration pipeline uses pre-trained models to predict both hand configurations and MCP joint angles based on forearm ultrasound images. First, a forearm ultrasound image is loaded. Then, a pre-trained model classifies the data into one of the several hand configuration classes. Based on this classification, another pre-trained model corresponding to the hand configuration is loaded. This model makes a prediction based on the loaded ultrasound image frame. The index, middle, ring, and pinky finger MCP angle prediction plot is then displayed. This is done repeatedly for a given number of ultrasound data frames which can be specified by the user to produce the output for the combined pipeline shown in Fig. \ref{fig:fig_1}.
\subsection{Hand Configuration Classification}
For hand configuration classification based on ultrasound images from the forearm, SVC with different kernels, a multi-layer perceptron (MLP) network, and a convolutional neural network (CNN) were used. The SVC models were trained for three different kernels, namely linear (SVC-Lin), radial basis function (SVC-RBF), and third-degree polynomial (SVC-Ply). For each subject and speed, 5.6 seconds of training data and 2.4 seconds of test data (70\% train-test split on 8 seconds of data) was acquired per hand configuration. For the 11 configurations, this lead to 88 seconds of data per speed per subject being used for training and testing the models. Different classifiers were used on full image sizes (636 x 256 pixels corresponding to 44.52 mm depth and 50 mm width forearm ultrasound images) and down-sampled image sizes (169 x 64 pixels, full images down-sampled by a factor of 4). The downsampling was done as a way to validate the performance of the classifiers at a lower resolution.
\subsubsection{Support Vector Classifier}
Support vector machines (SVMs) are a class of supervised machine learning algorithms that are used for image classification and regression problems. A support vector classifier (SVC) constructs hyperplanes in high-dimensional spaces, wherein a good separation is achieved by the hyperplane that has the largest distance to the nearest training-data point of any class. For ultrasound forearm image classification to predict hand configurations, SVC and other supervised machine learning algorithms have been explored in the past \cite{bimbraw2020towards, mcintosh2017echoflex}. Two-dimensional ultrasound images are the input to the algorithm, and the hand configuration class is the corresponding ground truth. After training the SVM model, ultrasound images not used for the training can be used to make a prediction of the corresponding hand configuration. 
\subsubsection{Multi Layer Perceptron}
An MLP is a fully connected feedforward neural network. A 5-layer MLP was used to classify the 11 hand configurations based on forearm ultrasound. The 2D images were first flattened into a vector and a Dense layer with a dimensionality of 128 and the ReLU activation function, the three successive layers  with dimensions 64, 32, and 16, with the same activation as the first hidden layer. The final layer had 11 elements and the Softmax activation function to make the prediction. 

\begin{figure}[t]
    \centering
    \includegraphics[scale=0.13]{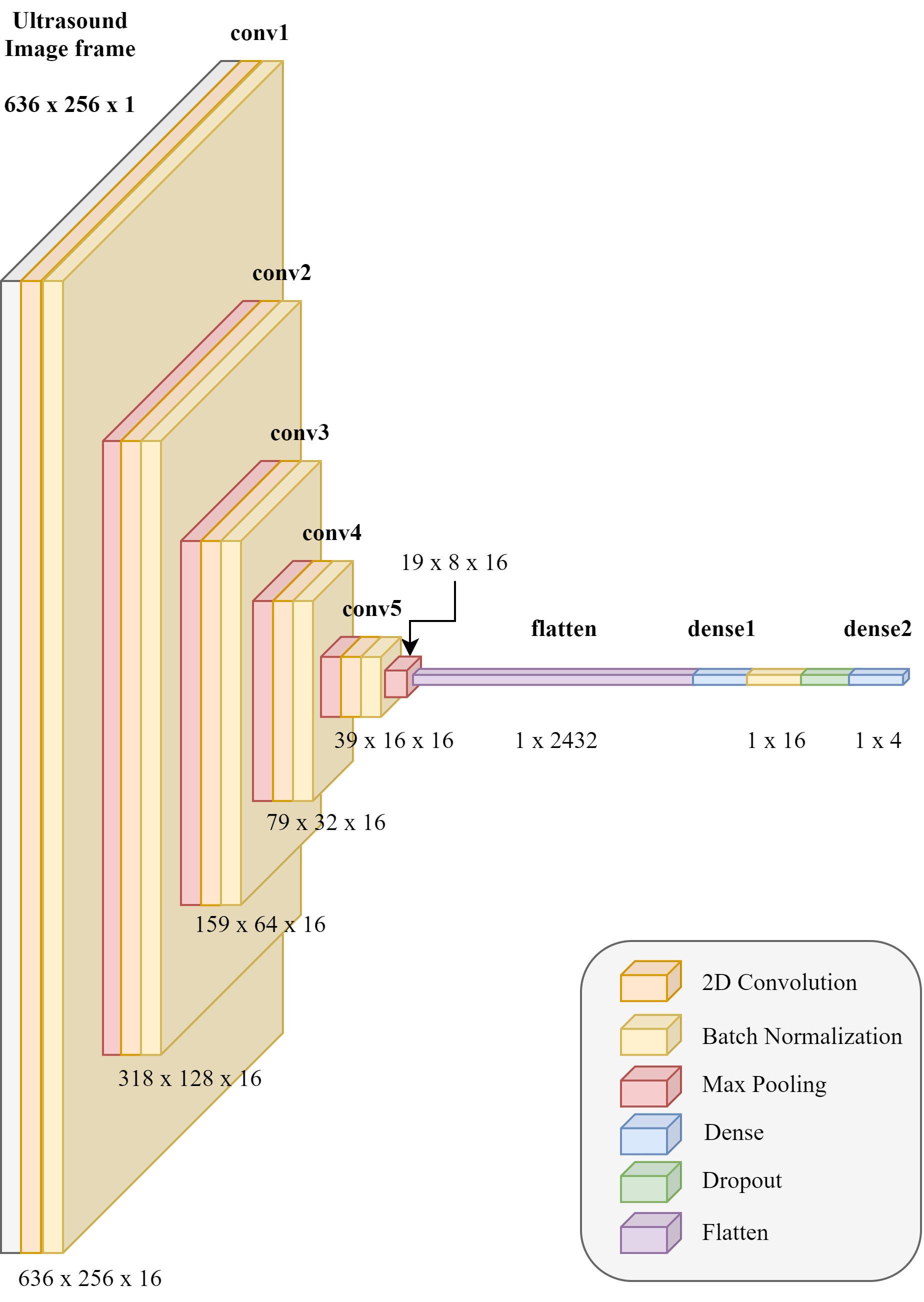}
    \caption{The CNN with 5 convolutional sections followed by flattening and then two dense layers which lead to the 1x4 output for predicting the finger angles for the index, middle, ring, and pinky fingers based on a single frame of a forearm ultrasound image.}
    \label{fig:fig_2}
\end{figure}

\subsubsection{Convolutional Neural Network}
The CNN architecture used for the classification is the same which was used for MCP joint angle estimation and is explained in section 2.C. The network can be visualized in Fig. \ref{fig:fig_2}.

\subsection{MCP joint angle estimation}
MCP joint angle prediction was implemented using a CNN. CNNs are a class of deep learning algorithms popularly used for two-dimensional image data regression-based prediction. The CNN architecture that was used in the paper is based on a modified form of VGG16 \cite{simonyan2014very}. The proposed CNN network has 5 convolutional layers each followed by batch normalization and a max pooling layer. This is different from the original VGG16 since it uses two and three cascaded convolutional layers for each of the 5 different convolutional sections of the network. Fig. \ref{fig:fig_2} shows the proposed CNN model.

The conv2d layer creates a convolution kernel that is convolved with the input layer to produce a tensor of outputs. The output \(Z_1\) after the convolution operation (\(*\)) is defined in \eqref{dl1}.
\begin{equation}Z_1 = X * f\label{dl1}\end{equation}
where \(X\) is the input image and \(f\) is the filter or the convolution kernel. For the proposed model, each conv2d layer has a kernel size of 3 x 3 pixels, with ReLU activation and padding with zeros evenly around the input. The ReLU activation function is defined in \cite{agarap2018deep}. The output \(A_1\) after the activation function \(relu\) is applied is defined in \eqref{dl2}.
\begin{equation}A_1 = relu(Z_1)\label{dl2}\end{equation}
The batch normalization layer normalizes its output using the mean and standard deviation of the current batch of inputs from the output of the previous layer \(A_1\). The batch normalization is defined in \cite{ioffe2015batch}. The output \(B_1\) after applying the batch normalization function \(bn\) to \(A_1\) is defined in \eqref{dl3}.
\begin{equation}B_1 = bn(A_1)\label{dl3}\end{equation}
The max pooling later downsamples the input along its spatial dimensions by taking the maximum value over an input window of size 2 x 2 pixels for each channel of the input. The window is shifted by strides of 1 pixel along each dimension. Max pooling operation is defined in \cite{ranzato2007unsupervised}. The output \(M_1\) after applying the max pooling function \(mp\) to \(B_1\) is defined in \eqref{dl4}. 
\begin{equation}M_1 = mp(B_1)\label{dl4}\end{equation}
This is repeated 5 times, with the successive convolution operations {\(Z_2\), \(Z_3\), \(Z_4\), \(Z_5\)}, successive ReLU activation function outputs {\(A_2\), \(A_3\), \(A_4\), \(A_5\)}, successive batch normalization function outputs {\(B_2\), \(B_3\), \(B_4\), \(B_5\)}, and successive max pooling function outputs {\(M_2\), \(M_3\), \(M_4\), \(M_5\)}. Therefore, the output \(O_1\) after all of these operations is defined in \eqref{dl5}. Each conv2d (\(*\))-\(relu\)-\(bn\)-\(mp\) pair is enclosed in curly brackets.
\begin{equation}
\label{dl5}
\begin{split}
O_1 = \{M_5(B_5(A_5(Z_5\{M_4(B_4(A_4(Z_4\{M_3(B_3\ldots\\
(A_3(Z_3\{M_2(B_2(A_2(Z_2\{M_1\})))\})))\})))\})))\}
\end{split}
\end{equation}
The flatten function \(fl\) reshapes the output of the previous max pooling layer \(O_1\) and reduces its dimension to the form of (\(1\)x\(n\)) where \(n\) is the product of the dimensions of the previous layer. The output \(FL\) after applying the flatten function \(fl\) is defined in \eqref{dl6}.
\begin{equation}FL = fl(O_1)\label{dl6}\end{equation}
The first dense layer is a deeply connected 16-unit neural network layer with ReLU activation. The output \(D_1\) of the densely connected neural network layer is defined in \eqref{dl7}.
\begin{equation}D_1 = relu(FL \cdot gu_f)\label{dl7}\end{equation}
where \(gu_f\) is the weights matrix created by the layer initialized by the Glorot/Xavier uniform initializer \cite{glorot2010understanding}. The ($\cdot$) denotes the dot product of the \(FL\) and the \(gu_f\). This is followed by batch normalization function \(bn\) with the output \(B_6\) defined in \eqref{dl8}.
\begin{equation}B_6 = bn(D_1)\label{dl8}\end{equation}
This is followed by the dropout function \(drop\) which randomly sets input units to 0 with a frequency of rate at each step during training time, which helps prevent overfitting \cite{srivastava2014dropout}. The output \(Dr_1\) is defined in \eqref{dl9}
\begin{equation}Dr_1 = drop(B_6)\label{dl9}\end{equation}
The final dense layer (the output \(Output\)) is a deeply connected neural network later with ReLU activation that generates the final predictions. This is defined in \eqref{dl10}.
\begin{equation}Output = relu(Dr_1 \cdot gu_f)\label{dl10}\end{equation}
After obtaining \(O_1\), the \(Output\) is defined in \eqref{dl11}.
\begin{equation}Output = relu\{drop(bn(relu\{fl(O_1) \cdot gu_f\}))\cdot gu_f\} \label{dl11}\end{equation}
The \(Output\) layer has 11 units for hand configuration classification and 4 units for MCP joint angle estimation. For training, forward propagation and backward propagation train an optimized model. For inference, forward propagation generates the output prediction. During forward propagation, the weights and filters are randomly initialized and are used as model parameters. During backward propagation, the model parameters are updated over successive epochs to reduce the loss and improve the accuracy.

\begin{figure}[t]
\centering
\includegraphics[scale=0.49]{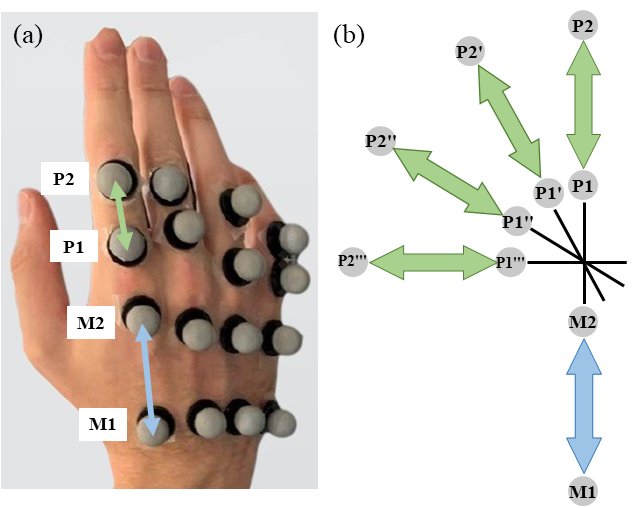}
\caption{(a) Location of markers and vectors on the hand, and (b) the definition for MCP angle estimation.}
\label{fig:Hand_Angle}
\end{figure}

\section{Experimental Implementation}
This section describes the data acquisition, hardware preparation, software used, and experimental testing protocol.  

\subsection{Human Subjects and IRB Approval}
The study was approved by the institutional research ethics committee at the Worcester Polytechnic Institute (No. IRB-21-0452), and written informed consent was given by the subjects prior to all sessions. Ultrasound data was captured from the right forearm of 6 subjects (1 female, 5 males; Age: 22.83 ± 3.14 years; Height: 173.55 ± 11.02 cm; diameter of the forearm around the point where the probe is placed: 20.43 ± 2.74 cm). Table \ref{tab:my_label} lists the age, sex, height, and forearm diameter at the ultrasound probe location for the 6 subjects enrolled in the study.

\begin{table}[t]
\centering
\caption{Subject Information}
\label{tab:my_label}
\begin{tabular}{|p{1.4cm}|p{0.58cm}|p{0.58cm}|p{0.58cm}|p{0.58cm}|p{0.58cm}|p{0.58cm}|}
\hline
& TS 1 & TS 2 & TS 3 & TS 4 & TS 5 & TS 6\\
\hline
Age (yr) & 25 & 20 & 24 & 23 & 24 & 21\\
Sex (M/F) & F & M & M & M & M & M\\
Height (cm) & 157.5 & 170.2 & 172.7 & 180.3 & 167.6 & 193.0\\
Forearm Dia. (cm) & 22.9 & 18.4 & 18.4 & 19.1 & 18.4 & 25.4\\
\hline
\end{tabular}
\end{table}

\subsection{Instrumentation}
Velcro straps on both sides of the custom-designed 3D-printed probe casing were used to strap the ultrasound probe on the forearm. The subject’s arm rests on a rest table which was further secured with another external Velcro extender to the table after verification of the ultrasound images. Fig. \ref{fig:components} shows a subject’s right arm strapped to the rest table.

\subsubsection{Hardware Preparation}
\begin{figure}[t]
    \centering
    \includegraphics[scale=0.105]{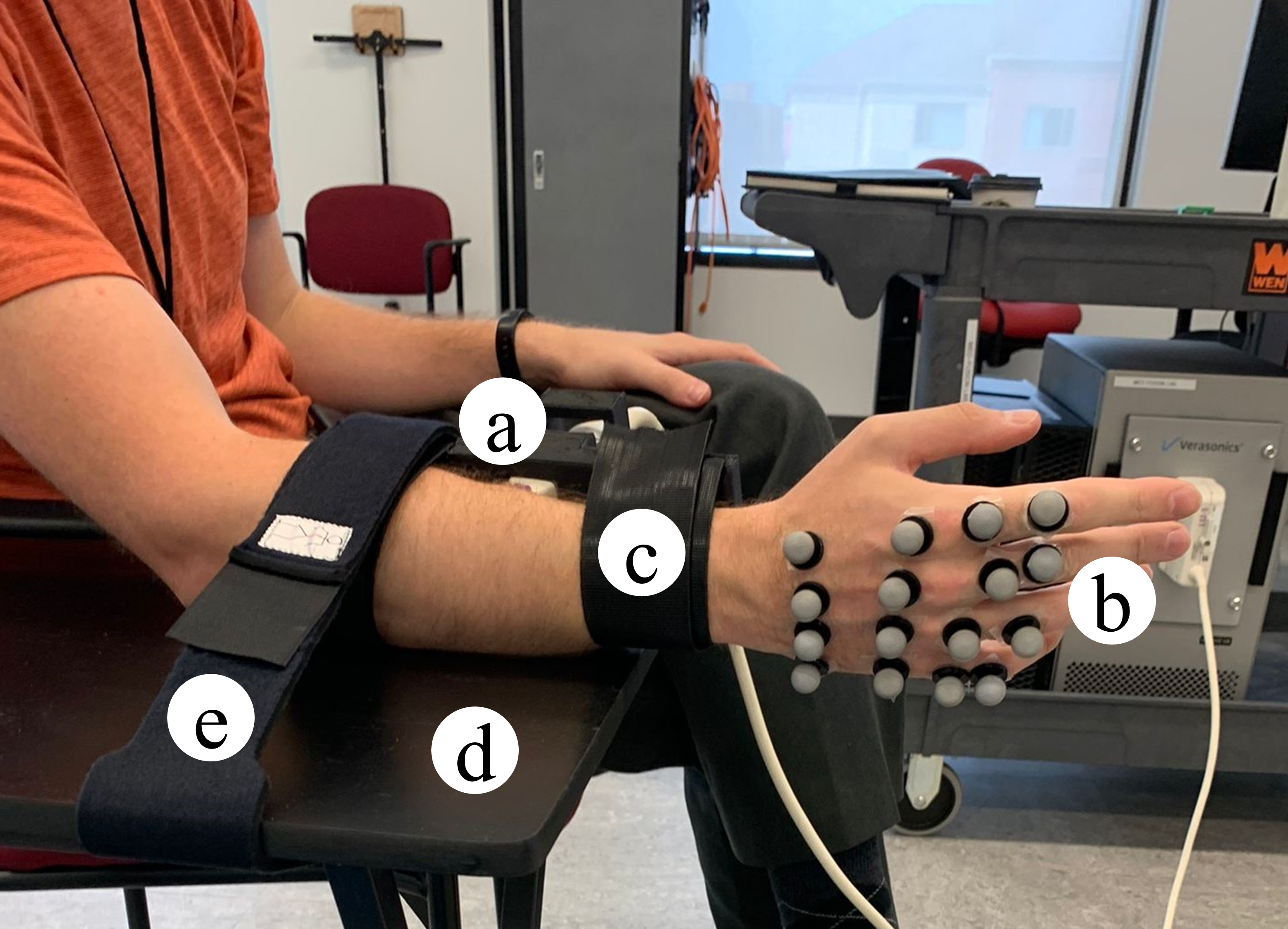}
    \caption{The ultrasound probe in a 3D printed casing (a) strapped on a subject’s right forearm (c) and the motion capture markers (b) on their right hand. The forearm is secured with a Velcro strap (e) around the rest table (d). The ultrasound probe is mounted transversally on the forearm.}
    \label{fig:components}
\end{figure}

Three-dimensional positional data of specified points on their index, middle, ring, and pinky fingers (Fig. \ref{fig:Hand_Angle}(a)) was acquired along with the ultrasound data, with the ultrasound probe mounted in a transverse position on the forearm (Fig. \ref{fig:components}). This data was acquired at three different speeds of hand movement. The experimental setup uses a Vantage 128 Verasonics ultrasound data acquisition system (Verasonics, WA, USA) and a Vicon Nexus motion capture system (Vicon Motion Systems Ltd., UK). The average estimated error linked to the MCP joint angle ground-truth estimation using the Vicon Nexus motion capture system has been reported to be less than the acceptable 5$^{\circ}$ except for the pinky MCP joint angle standard error of measurement which is less than 6$^{\circ}$ \cite{fischer2020development}.

An L12/5 50 mm linear array ultrasound probe was used with the Verasonics Vantage 128/128 research ultrasound system. A MATLAB script was used to set sequence objects for the Verasonics data acquisition hardware to display and record ultrasound images. A trigger output signal was sent to an Arduino board from the Verasonics system as a synchronizing signal. The ultrasound data frame acquisition rate was set at 25 Hz. For 56 seconds of data acquisition per session, 1400 frames were recorded. Each data frame was 636 x 256 pixels, corresponding to a 44.52 mm depth of the ultrasound image and a 50 mm width of the imaging region. 
\begin{figure}[t]
    \centering
    \includegraphics[scale=0.075]{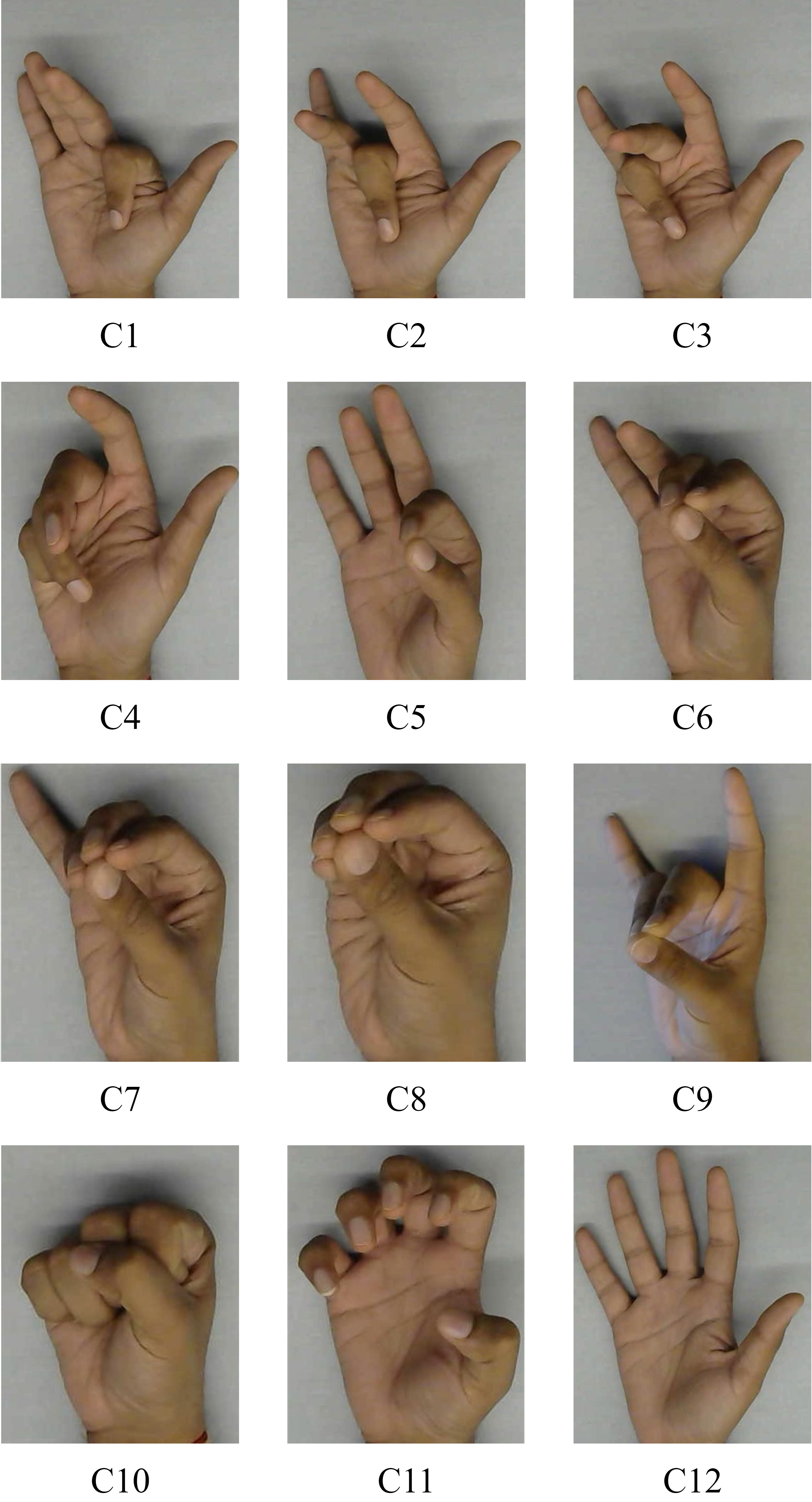}
    \caption{The hand configurations relevant to the activities of daily living (ADLs).}
    \label{fig:states}
\end{figure}
A 10-camera Vicon Vantage motion capture system with Lock+ 64-channel ADC (Analog to Digital Converter) was used to obtain the three-dimension (3-D) positional information for different motion capture markers attached to the finger. Because of a sample rate mismatch between the trigger from the Verasonics ultrasound system and the Vicon motion capture system, an Arduino Uno was used as a bridge to record trigger signals utilizing Arduino’s interrupt service routine and generate a pulse that could be read by the Vicon analog to digital converter.

\subsubsection{Audio Signal Generation}
Three audio waveforms were designed for simulating three different speeds for hand movements using Audacity, an open-source digital audio workstation. 0.1-second Sawtooth waves at 440 Hz and 392 Hz were used to alert the subjects to alternate between the rest and motion hand states. The frequencies of the rest/motion switching were set as 2 Hz for fast, 1 Hz for medium, and 0.5 Hz for slow speeds. These speeds were set so that in addition to the standard 1 Hz signal, data at speeds more and less than 1 Hz by a factor arbitrarily chosen as 2 could be quantitatively acquired.

\subsubsection{Computation Hardware and Software}
NVIDIA GeForce RTX 2070 SUPER was used to develop and debug the deep learning models. AMD Ryzen 7 2700X Eight-Core Processor was used to develop and debug the classical machine learning models. The system had 31.91 GB of available RAM. The code was executed in Python 3.7. TensorFlow Keras API was used for deep learning models like our custom CNN and MLP\cite{geron2019hands}. Scikit-learn library was used for classical machine learning based classification models \cite{pedregosa2011scikit}. The scikit-learn function that was used to implement SVC used C-Support Vector Classification (C-SVC) implementation based on libsvm \cite{pedregosa2011scikit}. The multiclass support is handled according to a one-vs-one scheme, and the 11 classes are divided into 55 binary classification datasets. Each of these 55 binary classification models predicts one class and the model with the most predictions is predicted by the one-vs-one scheme.

\begin{figure}[t]
    \centering
    \includegraphics[scale=0.17]{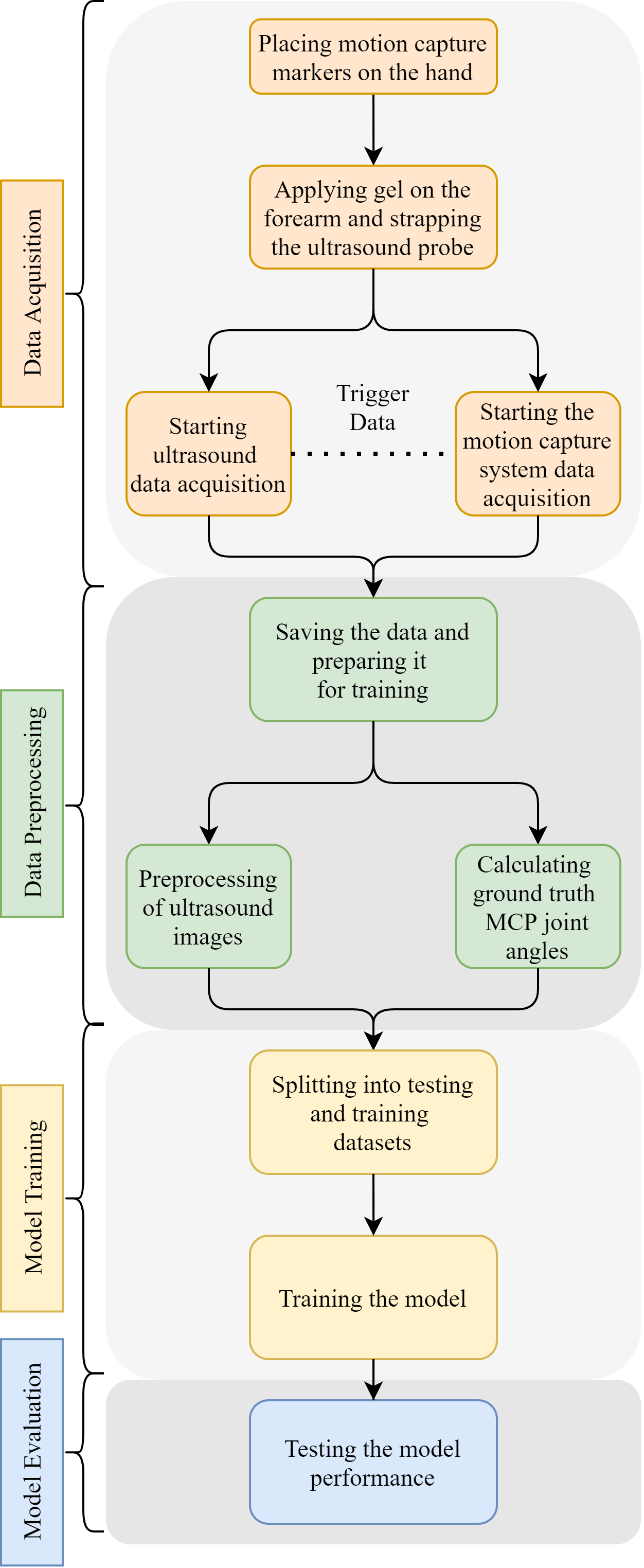}
    \caption{The data acquisition, processing, and evaluation pipeline consist of four components: (1) Ultrasound and motion capture data acquisition, (2) Data preprocessing to obtain ultrasound images and MCP joint angles as the ground truth, (3) Data splitting and model training, and (4) Model evaluation.}
    \label{fig:process1}
\end{figure}

\subsection{Experimental Testing Protocol}
11 hand configurations relevant to the activities of daily living (ADLs) were chosen for this project \cite{dollar2014classifying}. The hand configurations are shown in Fig. \ref{fig:states}. The hand configurations: C1 (IndFlex): Index finger flexion; C2 (MidFlex): Middle finger flexion; C3 (RinFlex): Ring finger flexion; C4 (PinFlex): Pinky finger flexion; C5 (IndPinch): Index finger in contact with the thumb; C6 (IndMidPinch): Index \& Middle fingers in contact with the thumb; C7 (IndMidRinPinch): Index, Middle \& Ring fingers in contact with the thumb; C8 (AllPinch): All fingertips touching; C9 (MidRinPinch): Middle \& Ring fingers in contact with the thumb; C10 (Fist); C11 (Hook): Movement is restrained to just the interphalangeal joint movement with the MCP joint angle remaining constant; Open: All fingers extended. The Open (all fingers extended) was considered as the rest state for all the 11 hand configurations and the subjects alternated between the open hand and selected the hand configuration. The subject was seated, and 16 motion capture markers were attached to the ends of the metacarpal and proximal phalanx for each of the index, middle, ring, and pinky fingers on their right hand as seen in Fig. \ref{fig:Hand_Angle} (a).

This marker configuration allowed us to calculate the MCP joint angle while ensuring free movement of the hand for the desired hand movements. The MCP joint angle was measured by taking the inverse cosine of the two vectors formed by each metacarpal-proximal phalanx pair. Ultrasound gel was applied to the subject’s forearm and the imaging surface on the ultrasound probe. The ultrasound probe was then strapped to the subject’s right forearm. Then, the subject’s forearm was fastened to a rest table. The subject then performed a trial run for the 11 hand configurations. Auditory beeps were used to indicate the desired rate of hand opening and closing. There was a rest of 1 minute between each data acquisition session and 2 minutes between each speed transition. 56 seconds of data were acquired for each data acquisition session. Per subject, there were 33 data acquisition sessions, for 11 different hand movements at 3 different speeds each.

\begin{figure*}[t]
\centering
\includegraphics[scale=0.17]{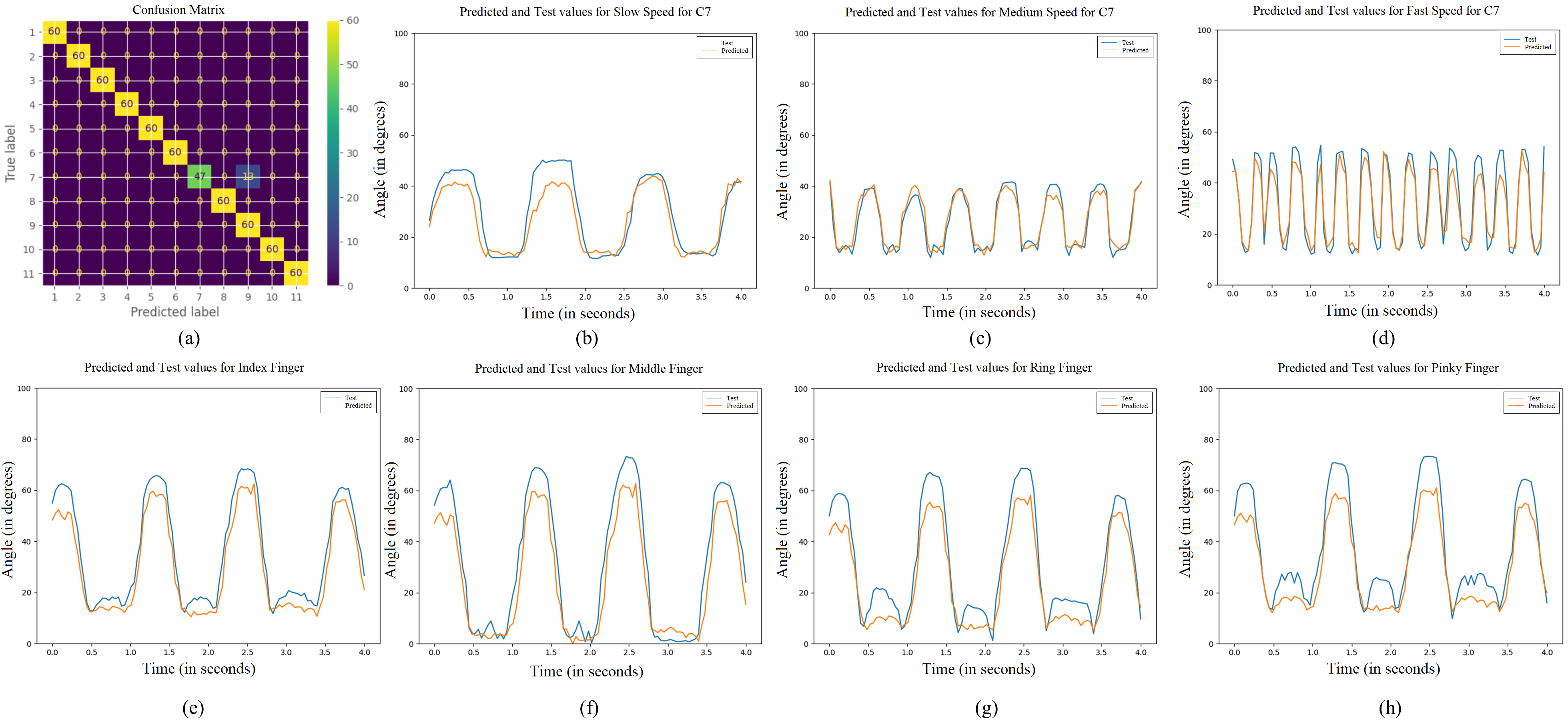}
\caption{(a) Confusion matrix for 11-state (C1 - C11: 0 - 10) classification for slow speed using the CNN. (b-d) For IndMidRinPinch, index finger predictions at slow (b), medium (c), and fast (d) speeds. (e-h) For fist at medium speed, the index (e), middle (f), ring (g), and pinky (h) finger predictions. (a-h) are for data obtained from Subject 1, and the colors blue and orange represent the test and predicted values respectively.}
\label{fig:results}
\end{figure*}

\section{Data Analysis}
Post-beamformed envelop detected ultrasound data were preprocessed before training the models by normalizing, log-compressing, and reshaping to highlight relevant muscle features. For the 6 subjects, data for 11 hand configurations, each at 3 speeds was acquired, leading to a total of 198 sessions of data acquisition. 56 seconds of data were acquired per subject, hand configuration, and speed, totaling 11088 seconds (6 * 11 * 3 * 56 seconds) of data used for analysis.

\subsection{Data and Model Parameters}
For the classification of hand configurations, small portions of each hand configuration for a specific subject and speed were used to validate the classification performance. For estimating the finger angles during the hand configuration motions, full data sizes for a specific subject, speed, and hand configuration were used to validate the regression performance.
\subsubsection{Classification Parameters}
For each subject for each hand configuration and speed, 8 seconds of data were considered. These files were then split with a train-test split of 70\% to get independent testing and training sets. 5.6 seconds of training and 2.4 seconds of test data per subject, hand configuration, and speed were used for generating the classification results. This data was then combined across the hand configurations for each subject and speed (88 seconds of data per subject and speed) leading to 61.6 seconds of training and 26.4 seconds of test data per subject and speed. 

\subsubsection{Regression Parameters}
The CNNs were trained for each hand configuration and speed on 56 seconds of data for every subject. Adam optimizer function was used as the gradient descent method. A learning rate of 1e-3 was set for the Adam optimizer \cite{kingma2014adam}. The learning rate decay was set to 1e-3/200. Mean Absolute Error was used as the loss function. It computes the mean of the absolute difference between true and predicted values. Each data file was split with a train-test split of 70\%. While training, the validation split was defined as 0.1. The number of epochs was set to 50.  

\subsection{Angle Estimation for Ground Truth}
Motion capture markers were placed on the subject's hand at the base and the head of the index, middle, ring, and pinky proximal phalanges and the metacarpals. The marker locations on a subject's right hand can be seen in Fig. \ref{fig:Hand_Angle}(a). Data was not acquired from the thumb because there is an extrinsic flexor for the interphalangeal joint in the forearm, but the more relevant motions of the thumb come from the carpometacarpal (CMC) joint and MCP joint. The motions of the CMC and MCP joints for the thumb depend on the intrinsic muscles that are not imaged. Hence, including the thumb angle, data will be the subject of future research. 

For the index metacarpal, the points can be defined as \(M1\) and \(M2\), with the vector $\overrightarrow{M1 M2}$, shown in red in Fig. \ref{fig:Hand_Angle}(a). For the index proximal phalanx, the points can be defined as \(P1\) and \(P2\), with the vector $\overrightarrow{P1 P2}$, shown in blue in Fig. \ref{fig:Hand_Angle}(a). The angle between the $\overrightarrow{M1 M2}$ and $\overrightarrow{P1 P2}$ is 180$^{\circ}$ measured counterclockwise. As seen in Fig. \ref{fig:Hand_Angle}(b), when the proximal phalanx flexes by 30$^{\circ}$ from its initial state ($\overrightarrow{P1 P2}$), the angle between the $\overrightarrow{M1 M2}$ and $\overrightarrow{P1' P2'}$ is 210$^{\circ}$. Similarly, when the proximal phalanx flexes by 30$^{\circ}$ and 60$^{\circ}$ from its initial state ($\overrightarrow{P1' P2'}$), the angles between the $\overrightarrow{M1 M2}$ - $\overrightarrow{P1'' P2''}$, and $\overrightarrow{M1 M2}$ - $\overrightarrow{P1''' P2'''}$ pairs are 240$^{\circ}$ and 270$^{\circ}$, respectively. The MCP joint angle was measured by taking the inverse cosine of the two vectors formed by each metacarpal-proximal phalanx pair.

\subsection{Quantification Metrics}
Accuracy percentage \(Acc\) was used for evaluating the classification performance. It is defined in \eqref{eq1}.
\begin{equation}Acc = \frac{TP + TN}{N} * 100\label{eq1}\end{equation}

\noindent where, \(TP\) is the number of True Positives, \(TN\) is the number of True Negatives, and \(N\) is the Total Sample Size. \(TP\) is an outcome where the model correctly predicts the positive class. \(TN\) is an outcome where the model correctly predicts the negative class. The classification error percentage is obtained by subtracting \(Acc\) from 100.

The Root Mean Squared Error (\(RMSE\)) values were used to quantify the regression results, defined in \eqref{eq2}.
\begin{equation}RMSE = \sqrt{ \frac{1}{N}\sum_{i=1}^{N} (y_{i}-\hat{y_{i}})^2}\label{eq2}\end{equation}

\noindent where, \(N\) is the Total Sample Size, \(i\) is the integer value ranging from 1 to the total number of samples, \(y_{i}\) is the test data value for the sample \(i\), and \(\hat{y_{i}}\) is the value predicted by the deep learning algorithm at the sample \(i\).

For quantifying both \(Acc\) and \(RMSE\), Arithmetic Mean (\(\mu\)) and Standard Deviation (\(\sigma\)) were used. Arithmetic mean is defined in \eqref{eq3}.
\begin{equation}\mu = \frac{1}{N}\sum_{i=1}^{N} a_{i}\label{eq3}\end{equation}

\noindent where, \(\mu\) is the arithmetic mean, \(N\) is the number of values, and \(a_{i}\) is the value from a data set of values, which for this paper can be \(Acc\) values or \(RMSE\) values. Standard deviation (\(\sigma\)) is defined in \eqref{eq4}.
\begin{equation}\sigma = \sqrt{ \frac{1}{N}\sum_{i=1}^{N} (x_{i}-\mu)}\label{eq4}\end{equation}

\noindent where, \(\sigma\) is the standard deviation, \(N\) is the number of values, and \(x_{i}\) is the value from a data set of values, and \(\mu\) is the mean of the data calculated in \eqref{eq3}.

\section{Results}
Here, the hand configuration classification results using SVC, MLP, and the proposed CNN, and MCP joint angle estimation results using the proposed CNN are described. A low latency MCP joint angle prediction and hand configuration classification pipeline aimed at real-time control of interfaces is also described. The results based on data obtained from subject number 1 are shown in Fig. \ref{fig:results}.

\subsection{Hand Configuration Classification}
To prove that the classifier can classify the 11 classes described in Fig. \ref{fig:states} based on the ultrasound images from the forearm, different machine learning models were trained for each subject for all the speeds. Fig. \ref{fig:results}(a) shows the confusion matrix using the CNN for slow-speed data from Subject 1.

\subsubsection{Hand configuration classification over different machine learning algorithms}
The highest classification accuracy was found for the CNN. Among the different SVC kernels, the polynomial kernel performed the best classification accuracy versus the linear and the radial basis function kernels whose performance was approximately similar. MLP had the least classification accuracy. The results are summarised in Table \ref{tab:5_A_1}.

\begin{table}[h]
\centering
\caption{Classification Performance over Different ML Algorithms}
\label{tab:5_A_1}
\begin{tabular}{|p{0.75cm}|p{1.25cm}|p{1.25cm}|p{1.25cm}|p{.75cm}|p{.75cm}|}
\hline
Metric & SVC-Lin & SVC-RBF & SVC-Ply & MLP & CNN\\
\hline
\(Acc\) & 80.6 & 80.2 & 81.5 & 72.6 & 82.7\\
\(\sigma\) & 10.7 & 10.8 & 10.9 & 9.3 & 9.7\\
\hline
\end{tabular}
\end{table}

\begin{figure}[b]
\centering
\includegraphics[scale=0.5]{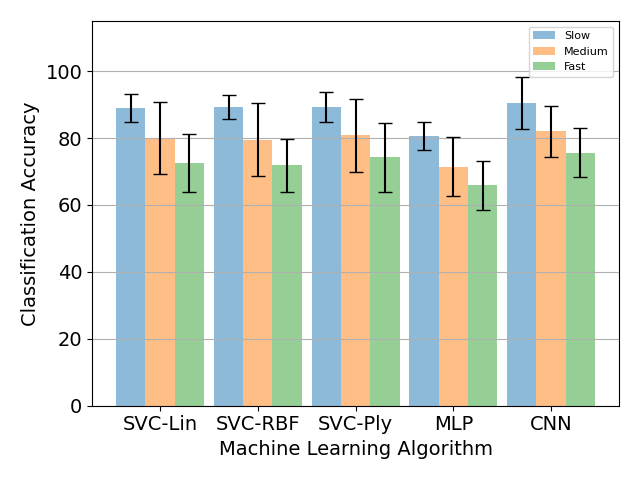}
\caption{Accuracy percentages for different speeds for the five different models for full images.}
\label{fig:5_a_2}
\end{figure}

\subsubsection{Hand configuration classification over different speeds}
For the five different models described in section 5.A.1, it was observed that the accuracy percentage was the highest for the Slow speed (0.5 Hz) and the lowest for the Fast speed (2 Hz). The consistency of this trend across the different algorithms is indicative of the gradation between the different speeds, with the fast speed images having more variability across subjects than the medium speed images, and the medium speed images having more variability than the slow speed images, which have less variability per subject across subjects among all the speeds, thereby leading to better performance. Like the previous section, the CNN performed the best with the slow speed accuracy of 90.5 $\pm$ 7.8\%, medium speed accuracy of 82.0 $\pm$ 7.5\%, and fast speed accuracy of 75.7 $\pm$ 7.3\%. Among the different SVC kernels, the polynomial kernel performed the best with the accuracy percentages of 89.3 $\pm$ 4.5\% for slow speed, 80.8 $\pm$ 10.9\% for medium speed, and 74.3 $\pm$ 10.3\% for fast speed versus the linear kernel (89.0 $\pm$ 4.2\% for slow speed, 80.0 $\pm$ 10.8\% for medium speed, 72.7 $\pm$ 8.6\% for fast speed) and the radial basis function kernel (89.3 $\pm$ 3.6\% for slow speed, 79.5 $\pm$ 11.0\% for medium speed, 71.8 $\pm$ 8.0\% for fast speed) whose performance was approximately similar. MLP performed the worst with the fast speed accuracy of 80.5 $\pm$ 4.2\%, medium speed accuracy of 71.5 $\pm$ 8.7\%, and fast speed accuracy of 65.8 $\pm$ 7.4\%. The results are summarised in Fig. \ref{fig:5_a_2}.

\subsubsection{Hand configuration classification over Down-Sampled image sizes}
For images down-sampled by a factor of 4, the highest classification accuracy was found for the SVC with the Polynomial Kernel. This was followed by the SVC with the linear and radial basis function kernels. Among the MLP and CNN, MLP delivered a better performance. The results are summarised in Table \ref{tab:5_A_3_1}.

\begin{table}[h]
\centering
\caption{Classification Performance over Different ML Algorithms for Down-Sampled Images}
\label{tab:5_A_3_1}
\begin{tabular}{|p{0.7cm}|p{1.2cm}|p{1.2cm}|p{1.2cm}|p{.75cm}|p{.75cm}|}
\hline
Metric & SVC-Lin & SVC-RBF & SVC-Ply & MLP & CNN\\
\hline
\(Acc\) & 80.6 & 80.1 & 81.8 & 79.6 & 79.3\\
\(\sigma\) & 4.0 & 3.8 & 4.1 & 4.1 & 3.7\\
\hline
\end{tabular}
\end{table}

\begin{figure}[b]
\centering
\includegraphics[scale=0.52]{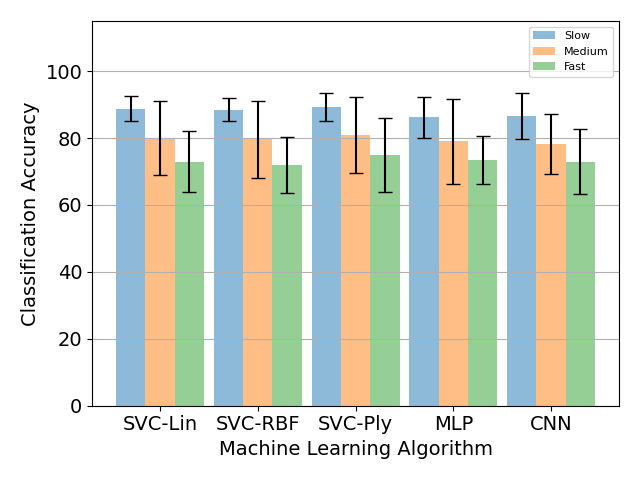}
\caption{Accuracy percentages for different hand configurations for the five different models for down-sampled images.}
\label{fig:5_a_3_2}
\end{figure}

Like 5.A.2, it was observed that the accuracy percentage was the highest for the Slow speed (0.5 Hz) and the lowest for the Fast speed (2 Hz). The SVC with polynomial kernel performed the best with the fast speed accuracy of 89.3 $\pm$ 4.1\%, medium speed accuracy of 81.0 $\pm$ 11.4\%, and fast speed accuracy of 75.0 $\pm$ 11.0\%. This was followed by SVC with the linear kernel at 88.8 $\pm$ 3.8\% for slow speed, 80.0 $\pm$ 11.1\% for medium speed, and 73.0 $\pm$ 9.1\% for fast speed. SVC with RBF kernel had the lowest performance among the three SVC kernels at 88.5 $\pm$ 3.5\% for slow speed, 79.7 $\pm$ 11.5\% for medium speed, and 72.0 $\pm$ 8.3\% for fast speed.  and the radial basis function kernel (89.3 $\pm$ 3.6\% for slow speed, 79.5 $\pm$ 11.0\% for medium speed, 71.8 $\pm$ 8.0\% for fast speed) whose performance was approximately similar. The MLP and CNN performance were comparable, at 86.2 $\pm$ 6.2\% for slow speed, 79.0 $\pm$ 12.8\% for medium speed, and 73.5 $\pm$ 7.1\% for fast speed for MLP, and the slow speed accuracy of 86.7 $\pm$ 6.9\%, medium speed accuracy of 78.2 $\pm$ 8.9\% and fast speed accuracy of 73.0 $\pm$ 9.7\%. The results are summarised in Fig. \ref{fig:5_a_3_2}.

\subsubsection{Classification Performance Comparison over Full and Down-Sampled Image Sizes}
\begin{figure}[h]
\centering
\includegraphics[scale=0.5]{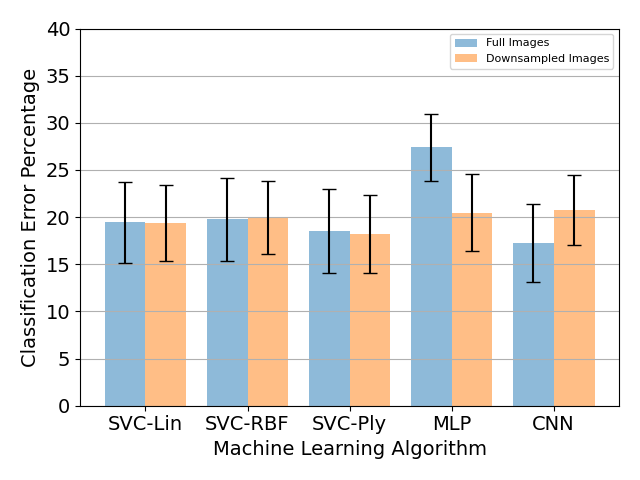}
\caption{Classification Error Percentage for different hand configurations for the five different models for full and down-sampled images.}
\label{fig:5_a_4}
\end{figure}

Among all the different algorithms, the highest classification percentage was obtained for CNN with full image data size at 82.7 $\pm$ 4.1\%. There was only a slight difference in accuracy percentage for SVC with linear, RBF, and polynomial kernels. For SVC with polynomial kernel and MLP, the results for down-sampled data were better than the full image data. But, these results were still worse than the CNN with full image sizes. Because the CNN results on full image sizes outperformed all other configurations, the Angle Estimation analysis was performed on the full image sizes using CNN described in Section 2.C. The comparison between full image sizes and down-sampled images is shown in Fig. \ref{fig:5_a_4}.

\subsection{Angle Estimation}
Using the CNN described in Section 2.C, CNN models were trained for each hand configuration, for each speed and subject (total of 6 * 11 * 3 = 198 models). Utilizing the CNN, a good degree of visual correspondence between the test data and the model predictions on individual trained models was obtained. This is shown for four second test data in Fig. \ref{fig:results}(b) through Fig. \ref{fig:results}(h). RMSE values were obtained from each of the hand configurations for each speed, for every subject. For the results, the full range of hand motion ranges from 0$^{\circ}$- 100$^{\circ}$. The range is set from 0$^{\circ}$-20$^{\circ}$ to highlight the averaged results and their gradation over the subjects and hand configurations. For models with successful convergence, the overall RMSE value averaged over all the hand configurations, three different speeds, and all the subjects was found to be 7.35 $\pm$ 1.3$^{\circ}$. These results show that by using the CNN architecture described in the paper, MCP joint angles can be predicted for different hand configurations based on ultrasound images from the forearm.

\subsubsection{Estimation Results for Different Hand Configurations}
To ensure the viability of the proposed approach for different hand configurations, RMSE values were calculated for the predictions averaged over all the subjects for all the slow, medium, and fast speeds for each hand configuration. Table \ref{tab:6_1} shows the plot of RMSE results for different hand configurations. The lowest average RMSE value was obtained for C11 (Hook) and the largest average RMSE value was obtained for C10 (Fist). The maximum standard deviation is observed for C10 (Fist) and the least standard deviation is observed for C11 (Hook). The high average RMSE and standard deviation for the fist can be explained by a big MCP joint change because of the movement of all the fingers during the motion. For hook, the MCP joint angle change is the least, and hence the average RMSE and the standard deviation are the least. 

\begin{table}[h]
\centering
\caption{Averaged \(RMSE\) for different Hand Configurations}
\label{tab:6_1}
\begin{tabular}{|p{0.8cm}|p{.25cm}|p{.25cm}|p{.25cm}|p{.25cm}|p{.25cm}|p{.25cm}|p{.25cm}|p{.25cm}|p{.25cm}|p{.28cm}|p{.25cm}|}
\hline
Metric & C1 & C2 & C3 & C4 & C5 & C6 & C7 & C8 & C9 & C10 & C11\\
\hline
\(RMSE\) & 7.7 & 8.0 & 8.8 & 9.7 & 4.9 & 6.1 & 6.7 & 6.7 & 9.1 & 11.4 & 4.5\\
\(\sigma\) & 2.1 & 2.8 & 2.5 & 4.7 & 1.4 & 1.5 & 2.4 & 1.5 & 3.7 & 4.9 & 1.3\\
\hline
\end{tabular}
\end{table}

\subsubsection{Repeatability over Subjects}
To ensure that the results are repeatable over the subjects, RMSE values were calculated for the predictions averaged over all the slow, medium, and fast speeds, and the hand configuration for each subject. Table \ref{tab:my_label} lists the age, sex, height, and forearm diameter at the ultrasound probe location for the 6 subjects enrolled in the study. Table \ref{tab:6_2} shows the plot of RMSE results for different subjects. The lowest average RMSE value was obtained from subject 4 and the largest average RMSE value was obtained from subject 3. The lowest standard deviation was obtained from subject 4 and the largest average standard deviation was obtained from subject 2.  

\begin{table}[h]
\centering
\caption{Averaged \(RMSE\) for different Subjects}
\label{tab:6_2}
\begin{tabular}{|p{0.8cm}|p{.6cm}|p{.6cm}|p{.6cm}|p{.6cm}|p{.6cm}|p{.6cm}|}
\hline
Metric & S 1 & S 2 & S 3 & S 4 & S 5 & S 6\\
\hline
\(RMSE\) &      6.1 &   8.6 &  9.2 &  5.6 & 7.9 & 6.7\\
\(\sigma\) &    2.3 &   3.9 &  3.3 &  1.7 & 3.1 & 1.7\\
\hline
\end{tabular}
\end{table}

\subsubsection{Estimation Results over Different Speeds}
\begin{figure}[h]
\centering
\includegraphics[scale=0.5]{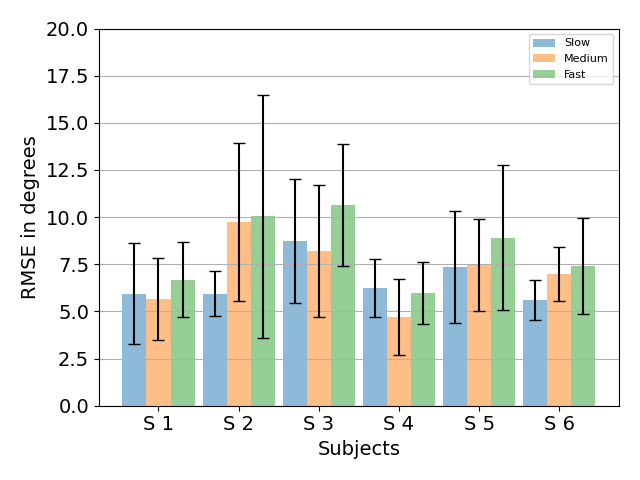}
\caption{RMSE results for subjects at different speeds averaged over the different configurations.}
\label{fig:5_b_3}
\end{figure}

Fig. \ref{fig:5_b_3} shows the RMSE results for subjects at different speeds averaged over the hand configurations. For subjects 2, 5, and 6, there is a trend of increasing RMSE as the hand movement speed increases from Slow (0.5 Hz) to Fast (2 Hz). For subjects 1, 3, and 4, the RMSE is higher for Slow speed (0.5 Hz) than for Medium speed (1 Hz). For all subjects except Subject 4, the RMSE is higher for Fast speed (2 Hz) than for Slow Speed (0.5 Hz). On averaging the speed over all subjects, it is observed that the average RMSE is the least for Slow speed at 6.67$^{\circ}$ $\pm$ 2.58$^{\circ}$. The average RMSE for Medium speed is 7.25$^{\circ}$ $\pm$ 3.32$^{\circ}$. The average RMSE is the highest for Fast speed at 8.36$^{\circ}$ $\pm$ 4.11$^{\circ}$. The standard deviation follows a similar trend and increases from Slow speed to Fast speed.

\subsection{Combined Prediction Pipeline}
For the combined prediction pipeline SVC-Lin (SVC with a Linear Kernel) models trained to classify between 11 different hand configurations for a single speed and subject and 11 CNN models trained for prediction of MCP joint angles were used. In the combined loop, first, the hand configuration state prediction is done using the saved SVC model. Then, based on the SVC prediction, the CNN model is chosen from a dictionary of saved models. For each frame of data, the SVC prediction takes 0.1 - 0.15 seconds. CNN prediction takes 0.004 - 0.007 seconds. The total time ranges from 0.11 - 0.16 seconds for angle and state prediction for one frame of ultrasound information, leading to 6.25 - 9.1 Hz of data processing capability using the combined prediction pipeline. The pipeline is described in Fig. \ref{fig:fig_1}.

\section{Discussion and Future Work}
The experiments and analysis show that both hand configuration classification and finger angle estimation can be obtained using one frame of ultrasound data using different classification and regression models. While ultrasound hand configuration classification is well researched, in this work, it is shown that with the experimental setup described in section 3, 82.7 $\pm$ 9.7\% classification accuracy with the CNN proposed in the paper and 81.5 $\pm$ 10.9\% classification accuracy using an SVC with a polynomial kernel was obtained. Due to computational limitations, classifiers that were based on ultrasound datasets larger than the ones used in the paper for classification were not used. It would be worthwhile to check out the classifier performance for bigger datasets and to see how deep learning based classification performs on ultrasound data in comparison with the traditional SVC. For real-time control applications, it would be necessary to evaluate the classification in an on-the-fly training and evaluation paradigm. Specifically, similar to the 4-state discretized classifier demonstrated in \cite{bimbraw2020towards}. The data was acquired in a way that motion artifacts were minimized, and the subjects were supervised in a way that they would be consistent with their movements for a particular hand configuration movement set which led to accuracy percentages over 80\% using both CNN and SVC with different kernels. It would be interesting to see the results in a system that is deployed for real-time use.

Several methods could be used for estimating hand movements such as a motion capture system \cite{yahya2019motion}, marker-less deep learning based hand movement detection sensors \cite{zhang2020mediapipe}, wearable gloves \cite{kessler1995evaluation}, etc. The motion capture system was used because of its ease of synchronization with the ultrasound data acquisition system and its wide use in literature for estimating finger joint angles \cite{ngeo2013control}. During the data acquisition, because of erratic hand movements and motion capture marker occlusion, there were issues with getting proper ground truth for some data acquisition sessions. 5.6\% of the total files couldn't be used because of this issue. An instrumented glove like the Cyberglove can be used as a future improvement to avoid issues related to markers not being visible to the camera in the motion capture system \cite{kessler1995evaluation}. Additionally, the proposed CNN model did not converge for 26.2\% of the files. Because of these issues, the MCP joint angle estimation results were obtained for 68.2\% of the total data files that were initially acquired. The model convergence issue for MCP joint angle estimation can be attributed to using one CNN model for all the subsets of the data over which the analysis was performed. Aggressive hyperparameter and model parameter optimization, along with increasing the model complexity for each of the subsets for which the model did not converge can help with obtaining convergence for more data. However, the updated parameters and hyperparameters might not work for the subsets for which the current model converges and thus not be generalizable over a more significant portion of the data, unlike the current model. There is ongoing research in the deep learning community on understanding generalization and its trade-off with model complexity \cite{zhang2021understanding}. Because of the good quality ultrasound data corresponding to the hand configurations, 100\% of the ultrasound data could be used for the classification. There were no issues with model convergence and the data labeling for the classification problem unlike the issues encountered with the MCP joint angle estimation problem. 

As can be seen in Table \ref{tab:6_1}, configurations C4 and C10 have the biggest RMSE values. This is because both these states (PinFlex and Fist) have significant pinky finger movement. Pinky finger movements were not consistent because it’s challenging to have repeatable movements with the pinky finger. It has been previously shown that the standard error of measurement among the MCP joint for the index, middle, ring, and pinky fingers is highest for the pinky finger \cite{fischer2020development}. This is reflected in our results since the hand movements primarily comprising the pinky finger MCP joint movement have the highest RMSE. For configuration C11 (Hook), the RMSE value was the minimum because the MCP angle values are smaller than other hand configurations. It can be seen that for the hand configurations involving lesser movements such as C5 (IndPinch) and C11 (Hook), the RMSE is lesser than the hand configurations involving more movement such as C4 (PinFlex) and C10 (Fist). The synergistic movements of multiple fingers in the hands and the corresponding effect on the RMSE needs to be investigated for a model that can be used to predict MCP joint angles regardless of the hand configuration. This would be a step in the direction of generalizability across hand configurations.

For the combined prediction pipeline, most of the time is taken by SVC prediction since the current implementation is not optimized on GPU and it uses the system CPU which is much slower. Exploring more GPU-supported deep neural network based classification algorithms can potentially improve the rate of simultaneous prediction from the current 6.25 - 9.1 Hz. In the current paper, different deep learning models were trained for each angle estimation data. It would be worthwhile to explore deep learning algorithms which can generalize the MCP joint angle prediction for different hand configurations as well as for different subjects. In the direction of portability, wearability, and technology translation, a system pipeline like the one demonstrated in \cite{bimbraw2020towards}, along with the low-latency deep learning based angle estimation system proposed in \cite{bimbraw2022prediction} and in this paper can be adapted for smaller and wearable ultrasound systems like the one demonstrated in \cite{wang2022bioadhesive} and other related works. The authors are excited about a future low-power (as the ones described in \cite{la2022flexible}), small form factor (like an Apple Watch), wearable (wearable hand gesture recognition interfaces surveyed in \cite{jiang2021emerging}), and portable device that utilizes forearm ultrasound data to estimate hand movements. This can be beneficial as a technology for human-machine interfacing and for deducing signals from the underlying musculature to develop rehabilitation tools \cite{patwardhan2022sonomyography}.

\section{Conclusions}
In this paper, the results for MCP joint angle estimation and hand state prediction based on forearm ultrasound are discussed. The proposed Convolutional Neural Network was able to get an average RMSE of 7.35$^{\circ}$ over 6 subjects, 3 speeds, and 11 hand configurations. Despite memory limitations, promising classification results were obtained on a subset of the entire dataset, with the proposed CNN leading to an average classification accuracy of 82.7\%, and SVC with linear, RBF, and polynomial kernels leading to an average classification accuracy over 80\%. A combined pipeline for angle and hand state estimation has also been described, which for the current setup can generate predictions as fast as 6.25 - 9.1 Hz per frame of ultrasound data. These results are encouraging, and they show the capability of ultrasound in understanding muscle movements. This work can inspire research in the promising domain of utilizing forearm ultrasound for predicting both continuous and discrete hand movements which can be useful for intuitive and adaptable control of physical robots and non-physical digital and AR/VR interfaces. 

\section*{Acknowledgment}

The authors would like to thank the members of the Medical Robotics Department at WPI for their input and help with the successful completion of the project. The authors are grateful to Tess Meier for her constructive criticism of the manuscript.

\bibliography{root.bib}

\begin{thebibliography}{10}

\bibitem{tarasenko2020artificial}
Anna Tarasenko, Mikheil Oganesyan, Daryna Ivaskevych, Sergii Tukaiev, Dauren
  Toleukhanov, and Nickolai Vysokov.
\newblock Artificial intelligence, brains, and beyond: Imperial college london
  neurotechnology symposium, 2020.
\newblock {\em Bioelectricity}, 2(3):310--313, 2020.

\bibitem{vuletic2019systematic}
Tijana Vuletic, Alex Duffy, Laura Hay, Chris McTeague, Gerard Campbell, and
  Madeleine Grealy.
\newblock Systematic literature review of hand gestures used in human computer
  interaction interfaces.
\newblock {\em International Journal of Human-Computer Studies}, 129:74--94,
  2019.

\bibitem{jung2021skin}
Yei~Hwan Jung, Jae-Hwan Kim, and John~A Rogers.
\newblock Skin-integrated vibrohaptic interfaces for virtual and augmented
  reality.
\newblock {\em Advanced Functional Materials}, 31(39):2008805, 2021.

\bibitem{zhu2020haptic}
Minglu Zhu, Zhongda Sun, Zixuan Zhang, Qiongfeng Shi, Tianyiyi He, Huicong Liu,
  Tao Chen, and Chengkuo Lee.
\newblock Haptic-feedback smart glove as a creative human-machine interface
  (hmi) for virtual/augmented reality applications.
\newblock {\em Science Advances}, 6(19):eaaz8693, 2020.

\bibitem{li2021survey}
Rui Li, Hongyu Wang, and Zhenyu Liu.
\newblock Survey on mapping human hand motion to robotic hands for
  teleoperation.
\newblock {\em IEEE Transactions on Circuits and Systems for Video Technology},
  2021.

\bibitem{makhataeva2020augmented}
Zhanat Makhataeva and Huseyin~Atakan Varol.
\newblock Augmented reality for robotics: A review.
\newblock {\em Robotics}, 9(2):21, 2020.

\bibitem{brown1994human}
Robert Brown and Charles~R Noback.
\newblock {\em Human anatomy \& physiology}.
\newblock McGraw-Hill, Incorporated, 1994.

\bibitem{cheok2019review}
Ming~Jin Cheok, Zaid Omar, and Mohamed~Hisham Jaward.
\newblock A review of hand gesture and sign language recognition techniques.
\newblock {\em International Journal of Machine Learning and Cybernetics},
  10(1):131--153, 2019.

\bibitem{ahsan2009emg}
Md~Rezwanul Ahsan, Muhammad~I Ibrahimy, Othman~O Khalifa, et~al.
\newblock Emg signal classification for human computer interaction: a review.
\newblock {\em European Journal of Scientific Research}, 33(3):480--501, 2009.

\bibitem{jiang2017exploration}
Xianta Jiang, Lukas-Karim Merhi, Zhen~Gang Xiao, and Carlo Menon.
\newblock Exploration of force myography and surface electromyography in hand
  gesture classification.
\newblock {\em Medical engineering \& physics}, 41:63--73, 2017.

\bibitem{oudah2020hand}
Munir Oudah, Ali Al-Naji, and Javaan Chahl.
\newblock Hand gesture recognition based on computer vision: a review of
  techniques.
\newblock {\em journal of Imaging}, 6(8):73, 2020.

\bibitem{guo2021human}
Lin Guo, Zongxing Lu, and Ligang Yao.
\newblock Human-machine interaction sensing technology based on hand gesture
  recognition: A review.
\newblock {\em IEEE Transactions on Human-Machine Systems}, 2021.

\bibitem{suarez2012hand}
Jesus Suarez and Robin~R Murphy.
\newblock Hand gesture recognition with depth images: A review.
\newblock In {\em 2012 IEEE RO-MAN: the 21st IEEE international symposium on
  robot and human interactive communication}, pages 411--417. IEEE, 2012.

\bibitem{ahmed2020device}
Hasmath Farhana~Thariq Ahmed, Hafisoh Ahmad, and CV~Aravind.
\newblock Device free human gesture recognition using wi-fi csi: A survey.
\newblock {\em Engineering Applications of Artificial Intelligence}, 87:103281,
  2020.

\bibitem{dipietro2008survey}
Laura Dipietro, Angelo~M Sabatini, and Paolo Dario.
\newblock A survey of glove-based systems and their applications.
\newblock {\em Ieee transactions on systems, man, and cybernetics, part c
  (applications and reviews)}, 38(4):461--482, 2008.

\bibitem{rautaray2015vision}
Siddharth~S Rautaray and Anupam Agrawal.
\newblock Vision based hand gesture recognition for human computer interaction:
  a survey.
\newblock {\em Artificial intelligence review}, 43(1):1--54, 2015.

\bibitem{chakraborty2018review}
Biplab~Ketan Chakraborty, Debajit Sarma, Manas~Kamal Bhuyan, and Karl~F
  MacDorman.
\newblock Review of constraints on vision-based gesture recognition for
  human--computer interaction.
\newblock {\em IET Computer Vision}, 12(1):3--15, 2018.

\bibitem{esposito2021biosignal}
Daniele Esposito, Jessica Centracchio, Emilio Andreozzi, Gaetano~D Gargiulo,
  Ganesh~R Naik, and Paolo Bifulco.
\newblock Biosignal-based human--machine interfaces for assistance and
  rehabilitation: A survey.
\newblock {\em Sensors}, 21(20):6863, 2021.

\bibitem{li2020review}
Kexiang Li, Jianhua Zhang, Lingfeng Wang, Minglu Zhang, Jiayi Li, and Shancheng
  Bao.
\newblock A review of the key technologies for semg-based human-robot
  interaction systems.
\newblock {\em Biomedical Signal Processing and Control}, 62:102074, 2020.

\bibitem{graupe1982multifunctional}
Daniel Graupe, Javad Salahi, and Kate~H Kohn.
\newblock Multifunctional prosthesis and orthosis control via microcomputer
  identification of temporal pattern differences in single-site myoelectric
  signals.
\newblock {\em Journal of Biomedical Engineering}, 4(1):17--22, 1982.

\bibitem{lee2020image}
Ung~Hee Lee, Justin Bi, Rishi Patel, David Fouhey, and Elliott Rouse.
\newblock Image transformation and cnns: A strategy for encoding human
  locomotor intent for autonomous wearable robots.
\newblock {\em IEEE Robotics and Automation Letters}, 5(4):5440--5447, 2020.

\bibitem{kahanowich2021robust}
Nadav~D Kahanowich and Avishai Sintov.
\newblock Robust classification of grasped objects in intuitive human-robot
  collaboration using a wearable force-myography device.
\newblock {\em IEEE Robotics and Automation Letters}, 6(2):1192--1199, 2021.

\bibitem{dong2021soft}
Wentao Dong, Lin Yang, Raffaele Gravina, and Giancarlo Fortino.
\newblock Soft wrist-worn multi-functional sensor array for real-time hand
  gesture recognition.
\newblock {\em IEEE Sensors Journal}, 2021.

\bibitem{rabe2020use}
Kaitlin~G Rabe, Mohammad~Hassan Jahanandish, Kenneth Hoyt, and Nicholas~P Fey.
\newblock Use of sonomyographic sensing to estimate knee angular velocity
  during varying modes of ambulation.
\newblock In {\em 2020 42nd Annual International Conference of the IEEE
  Engineering in Medicine \& Biology Society (EMBC)}, pages 3799--3802. IEEE,
  2020.

\bibitem{araki2012artificial}
Nozomu Araki, Kenji Inaya, Yasuo Konishi, and Kunihiko Mabuchi.
\newblock An artificial finger robot motion control based on finger joint angle
  estimation from emg signals for a robot prosthetic hand system.
\newblock In {\em The 2012 international conference on advanced mechatronic
  systems}, pages 109--111. IEEE, 2012.

\bibitem{shrirao2009neural}
Nikhil~A Shrirao, Narender~P Reddy, and Durga~R Kosuri.
\newblock Neural network committees for finger joint angle estimation from
  surface emg signals.
\newblock {\em Biomedical engineering online}, 8(1):1--11, 2009.

\bibitem{wang2020semg}
Chao Wang, Weiyu Guo, Hang Zhang, Linlin Guo, Changcheng Huang, and Chuang Lin.
\newblock semg-based continuous estimation of grasp movements by long-short
  term memory network.
\newblock {\em Biomedical Signal Processing and Control}, 59:101774, 2020.

\bibitem{xiong2021deep}
Dezhen Xiong, Daohui Zhang, Xingang Zhao, and Yiwen Zhao.
\newblock Deep learning for emg-based human-machine interaction: A review.
\newblock {\em IEEE/CAA Journal of Automatica Sinica}, 8(3):512--533, 2021.

\bibitem{tosin2022semg}
Maur{\'\i}cio~Cagliari Tosin, Juliano~Costa Machado, and Alexandre Balbinot.
\newblock semg-based upper limb movement classifier: Current scenario and
  upcoming challenges.
\newblock {\em Journal of Artificial Intelligence Research}, 75:83--127, 2022.

\bibitem{palumbo2021biopotential}
Arrigo Palumbo, Patrizia Vizza, Barbara Calabrese, and Nicola Ielpo.
\newblock Biopotential signal monitoring systems in rehabilitation: A review.
\newblock {\em Sensors}, 21(21):7172, 2021.

\bibitem{zhou2022non}
Hao Zhou and Gursel Alici.
\newblock Non-invasive human-machine interface (hmi) systems with hybrid
  on-body sensors for controlling upper-limb prosthesis: A review.
\newblock {\em IEEE Sensors Journal}, 2022.

\bibitem{ortega2018fatigue}
Pablo~A Ortega-Auriol, Thor~F Besier, Winston~D Byblow, and Angus~JC McMorland.
\newblock Fatigue influences the recruitment, but not structure, of muscle
  synergies.
\newblock {\em Frontiers in human neuroscience}, 12:217, 2018.

\bibitem{bimbraw2020towards}
Keshav Bimbraw, Elizabeth Fox, Gil Weinberg, and Frank~L Hammond.
\newblock Towards sonomyography-based real-time control of powered prosthesis
  grasp synergies.
\newblock In {\em 2020 42nd Annual International Conference of the IEEE
  Engineering in Medicine \& Biology Society (EMBC)}, pages 4753--4757. IEEE,
  2020.

\bibitem{zheng2006sonomyography}
Yong-Ping Zheng, MMF Chan, Jun Shi, Xin Chen, and Qing-Hua Huang.
\newblock Sonomyography: Monitoring morphological changes of forearm muscles in
  actions with the feasibility for the control of powered prosthesis.
\newblock {\em Medical engineering \& physics}, 28(5):405--415, 2006.

\bibitem{chen2010sonomyography}
Xin Chen, Yong-Ping Zheng, Jing-Yi Guo, and Jun Shi.
\newblock Sonomyography (smg) control for powered prosthetic hand: a study with
  normal subjects.
\newblock {\em Ultrasound in medicine \& biology}, 36(7):1076--1088, 2010.

\bibitem{shi2010feasibility}
Jun Shi, Qian Chang, and Yong-Ping Zheng.
\newblock Feasibility of controlling prosthetic hand using sonomyography signal
  in real time: preliminary study.
\newblock {\em Journal of Rehabilitation Research \& Development}, 47(2), 2010.

\bibitem{akhlaghi2015real}
Nima Akhlaghi, Clayton~A Baker, Mohamed Lahlou, Hozaifah Zafar, Karthik~G
  Murthy, Huzefa~S Rangwala, Jana Kosecka, Wilsaan~M Joiner, Joseph~J
  Pancrazio, and Siddhartha Sikdar.
\newblock Real-time classification of hand motions using ultrasound imaging of
  forearm muscles.
\newblock {\em IEEE Transactions on Biomedical Engineering}, 63(8):1687--1698,
  2015.

\bibitem{yang2020wearable}
Xingchen Yang, Yu~Zhou, and Honghai Liu.
\newblock Wearable ultrasound-based decoding of simultaneous wrist/hand
  kinematics.
\newblock {\em IEEE Transactions on Industrial Electronics}, 2020.

\bibitem{mcintosh2017echoflex}
Jess McIntosh, Asier Marzo, Mike Fraser, and Carol Phillips.
\newblock Echoflex: Hand gesture recognition using ultrasound imaging.
\newblock In {\em Proceedings of the 2017 CHI Conference on Human Factors in
  Computing Systems}, pages 1923--1934, 2017.

\bibitem{huang2017ultrasound}
Youjia Huang, Xingchen Yang, Yuefeng Li, Dalin Zhou, Keshi He, and Honghai Liu.
\newblock Ultrasound-based sensing models for finger motion classification.
\newblock {\em IEEE journal of biomedical and health informatics},
  22(5):1395--1405, 2017.

\bibitem{gerlich2007gesture}
L~Gerlich, Bernard~N Parsons, Anthony~S White, S~Prior, and Peter Warner.
\newblock Gesture recognition for control of rehabilitation robots.
\newblock {\em Cognition, Technology \& Work}, 9(4):189--207, 2007.

\bibitem{gu2022review}
Yuexing Gu, Yuanjing Xu, Yuling Shen, Hanyu Huang, Tongyou Liu, Lei Jin, Hang
  Ren, and Jinwu Wang.
\newblock A review of hand function rehabilitation systems based on hand motion
  recognition devices and artificial intelligence.
\newblock {\em Brain Sciences}, 12(8):1079, 2022.

\bibitem{simonyan2014very}
Karen Simonyan and Andrew Zisserman.
\newblock Very deep convolutional networks for large-scale image recognition.
\newblock {\em arXiv preprint arXiv:1409.1556}, 2014.

\bibitem{agarap2018deep}
Abien~Fred Agarap.
\newblock Deep learning using rectified linear units (relu).
\newblock {\em arXiv preprint arXiv:1803.08375}, 2018.

\bibitem{ioffe2015batch}
Sergey Ioffe and Christian Szegedy.
\newblock Batch normalization: Accelerating deep network training by reducing
  internal covariate shift.
\newblock In {\em International conference on machine learning}, pages
  448--456. PMLR, 2015.

\bibitem{ranzato2007unsupervised}
Marc'Aurelio Ranzato, Fu~Jie Huang, Y-Lan Boureau, and Yann LeCun.
\newblock Unsupervised learning of invariant feature hierarchies with
  applications to object recognition.
\newblock In {\em 2007 IEEE conference on computer vision and pattern
  recognition}, pages 1--8. IEEE, 2007.

\bibitem{glorot2010understanding}
Xavier Glorot and Yoshua Bengio.
\newblock Understanding the difficulty of training deep feedforward neural
  networks.
\newblock In {\em Proceedings of the thirteenth international conference on
  artificial intelligence and statistics}, pages 249--256. JMLR Workshop and
  Conference Proceedings, 2010.

\bibitem{srivastava2014dropout}
Nitish Srivastava, Geoffrey Hinton, Alex Krizhevsky, Ilya Sutskever, and Ruslan
  Salakhutdinov.
\newblock Dropout: a simple way to prevent neural networks from overfitting.
\newblock {\em The journal of machine learning research}, 15(1):1929--1958,
  2014.

\bibitem{fischer2020development}
Gabriella Fischer, Diana Jermann, Renate List, Lisa Reissner, and Maurizio
  Calcagni.
\newblock Development and application of a motion analysis protocol for the
  kinematic evaluation of basic and functional hand and finger movements using
  motion capture in a clinical setting—a repeatability study.
\newblock {\em Applied Sciences}, 10(18):6436, 2020.

\bibitem{geron2019hands}
Aur{\'e}lien G{\'e}ron.
\newblock {\em Hands-on machine learning with Scikit-Learn, Keras, and
  TensorFlow: Concepts, tools, and techniques to build intelligent systems}.
\newblock O'Reilly Media, 2019.

\bibitem{pedregosa2011scikit}
Fabian Pedregosa, Ga{\"e}l Varoquaux, Alexandre Gramfort, Vincent Michel,
  Bertrand Thirion, Olivier Grisel, Mathieu Blondel, Peter Prettenhofer, Ron
  Weiss, Vincent Dubourg, et~al.
\newblock Scikit-learn: Machine learning in python.
\newblock {\em the Journal of machine Learning research}, 12:2825--2830, 2011.

\bibitem{dollar2014classifying}
Aaron~M Dollar.
\newblock Classifying human hand use and the activities of daily living.
\newblock In {\em The Human Hand as an Inspiration for Robot Hand Development},
  pages 201--216. Springer, 2014.

\bibitem{kingma2014adam}
Diederik~P Kingma and Jimmy Ba.
\newblock Adam: A method for stochastic optimization.
\newblock {\em arXiv preprint arXiv:1412.6980}, 2014.

\bibitem{yahya2019motion}
Muhammad Yahya, Jawad~Ali Shah, Kushsairy~Abdul Kadir, Zulkhairi~M Yusof,
  Sheroz Khan, and Arif Warsi.
\newblock Motion capture sensing techniques used in human upper limb motion: A
  review.
\newblock {\em Sensor Review}, 2019.

\bibitem{zhang2020mediapipe}
Fan Zhang, Valentin Bazarevsky, Andrey Vakunov, Andrei Tkachenka, George Sung,
  Chuo-Ling Chang, and Matthias Grundmann.
\newblock Mediapipe hands: On-device real-time hand tracking.
\newblock {\em arXiv preprint arXiv:2006.10214}, 2020.

\bibitem{kessler1995evaluation}
G~Drew Kessler, Larry~F Hodges, and Neff Walker.
\newblock Evaluation of the cyberglove as a whole-hand input device.
\newblock {\em ACM Transactions on Computer-Human Interaction (TOCHI)},
  2(4):263--283, 1995.

\bibitem{ngeo2013control}
Jimson Ngeo, Tomoya Tamei, Tomohiro Shibata, MF~Felix Orlando, Laxmidhar
  Behera, Anupam Saxena, and Ashish Dutta.
\newblock Control of an optimal finger exoskeleton based on continuous joint
  angle estimation from emg signals.
\newblock In {\em 2013 35th Annual International Conference of the IEEE
  Engineering in Medicine and Biology Society (EMBC)}, pages 338--341. IEEE,
  2013.

\bibitem{zhang2021understanding}
Chiyuan Zhang, Samy Bengio, Moritz Hardt, Benjamin Recht, and Oriol Vinyals.
\newblock Understanding deep learning (still) requires rethinking
  generalization.
\newblock {\em Communications of the ACM}, 64(3):107--115, 2021.

\bibitem{bimbraw2022prediction}
Keshav Bimbraw, Christopher~J Nycz, Matthew~J Schueler, Ziming Zhang, and
  Haichong~K Zhang.
\newblock Prediction of metacarpophalangeal joint angles and classification of
  hand configurations based on ultrasound imaging of the forearm.
\newblock In {\em 2022 International Conference on Robotics and Automation
  (ICRA)}, pages 91--97. IEEE, 2022.

\bibitem{wang2022bioadhesive}
Chonghe Wang, Xiaoyu Chen, Liu Wang, Mitsutoshi Makihata, Hsiao-Chuan Liu, Tao
  Zhou, and Xuanhe Zhao.
\newblock Bioadhesive ultrasound for long-term continuous imaging of diverse
  organs.
\newblock {\em Science}, 377(6605):517--523, 2022.

\bibitem{la2022flexible}
Thanh-Giang La and Lawrence~H Le.
\newblock Flexible and wearable ultrasound device for medical applications: A
  review on materials, structural designs, and current challenges.
\newblock {\em Advanced Materials Technologies}, 7(3):2100798, 2022.

\bibitem{jiang2021emerging}
Shuo Jiang, Peiqi Kang, Xinyu Song, Benny~PL Lo, and Peter~B Shull.
\newblock Emerging wearable interfaces and algorithms for hand gesture
  recognition: A survey.
\newblock {\em IEEE Reviews in Biomedical Engineering}, 15:85--102, 2021.

\bibitem{patwardhan2022sonomyography}
Shriniwas Patwardhan, Jonathon Schofield, Wilsaan~M Joiner, and Siddhartha
  Sikdar.
\newblock Sonomyography shows feasibility as a tool to quantify joint movement
  at the muscle level.
\newblock In {\em 2022 International Conference on Rehabilitation Robotics
  (ICORR)}, pages 1--5. IEEE, 2022.

\end{thebibliography}

\newpage

\section{Biography Section}
\vskip -2\baselineskip plus -1fil
\begin{IEEEbiography}[{\includegraphics[width=1in,height=1.25in,clip,keepaspectratio]{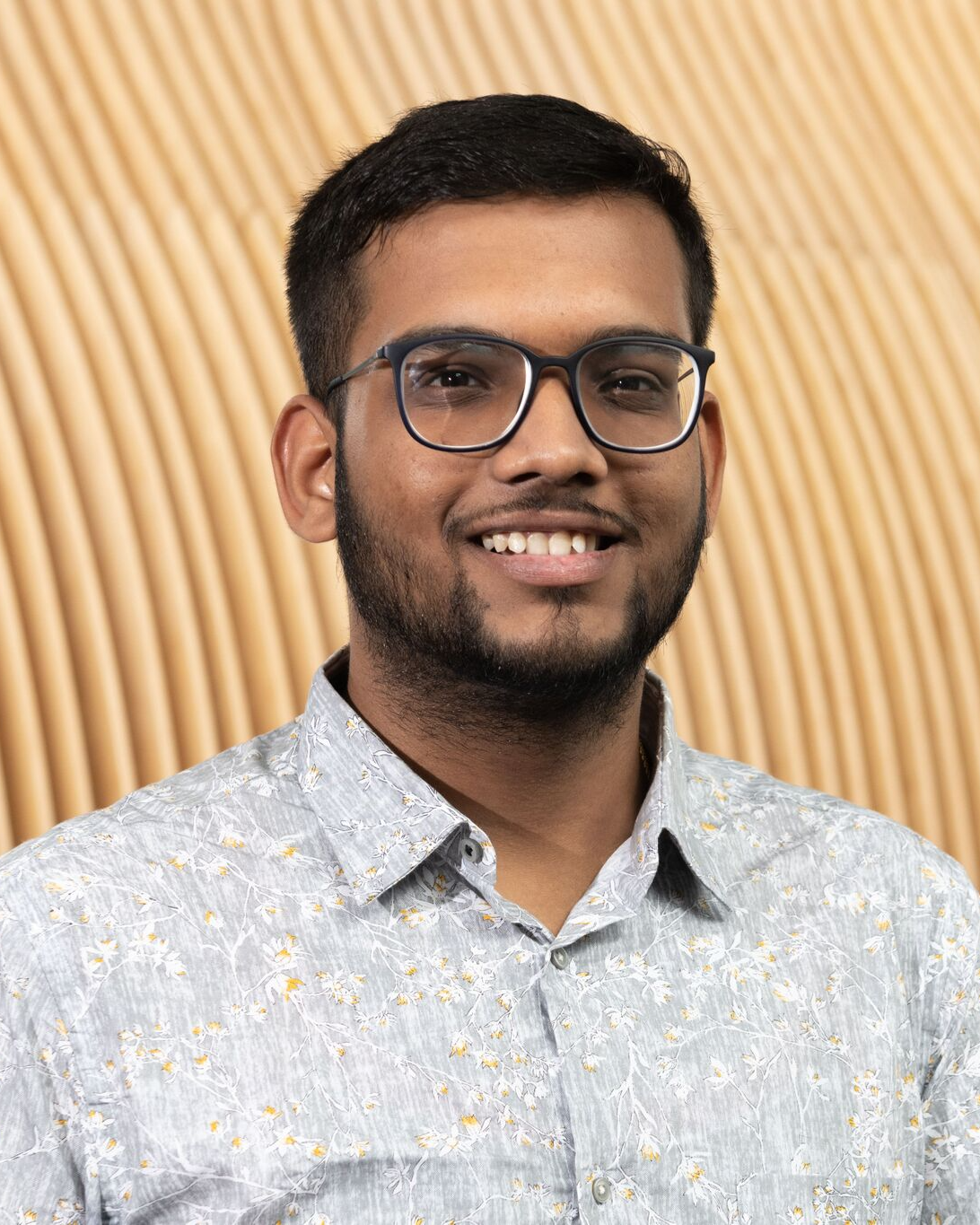}}]
{Keshav Bimbraw} is currently at Nokia Bell Labs as an Augmented Human Sensing Co-Op under Data and Devices Group in the Artificial Intelligence Research Lab. He is also a Ph.D. student in Robotics Engineering at Worcester Polytechnic Institute, 100 Institute Rd, Worcester, MA 01609, USA. He has an M. S. in Computer Software and Media Applications from Georgia Institute of Technology, Atlanta, GA 30332, USA, and a B. E. in Mechatronics Engineering from Thapar University, India. (e-mail: kbimbraw@wpi.edu, bimbrawkeshav@gmail.com).
\end{IEEEbiography}

\vskip -2\baselineskip plus -1fil

\begin{IEEEbiography}[{\includegraphics[width=1in,height=1.25in,clip,keepaspectratio]{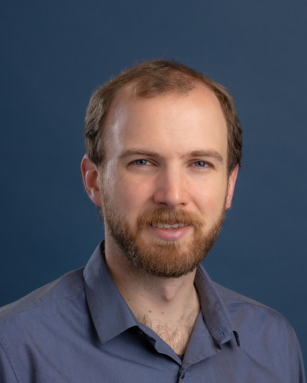}}]{Christopher J. Nycz} is a Research Scientist working with Worcester Polytechnic Institute's PracticePoint, 50 Prescott Street, Worcester, MA 01605, USA. He obtained Ph.D. and M. S. in Robotics Engineering from Worcester Polytechnic Institute, 100 Institute Rd, Worcester, MA 01609, USA. He holds a B. S. in Mechanical Engineering from Clarkson University, 8 Clarkson Ave, Potsdam, NY 13699, USA. (e-mail: cjnycz@wpi.edu)
\end{IEEEbiography}

\vskip -2\baselineskip plus -1fil

\begin{IEEEbiography}[{\includegraphics[width=1in,height=1.25in,clip,keepaspectratio]{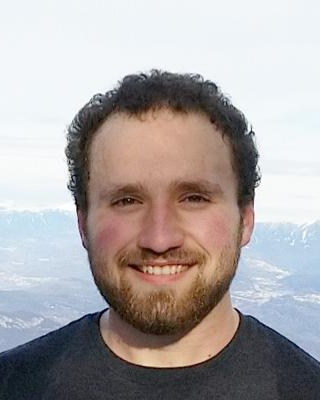}}]{Matthew Schueler} works as a Firmware Engineer at Amazon Robotics, 50 Otis St, Westborough, MA 01581, USA. He holds an M. S. Robotics Engineering and a B. S. in Robotics Engineering/Computer Science from Worcester Polytechnic Institute, 100 Institute Rd, Worcester, MA 01609, USA. (e-mail: mschueler@wpi.edu).
\end{IEEEbiography}

\vskip -2\baselineskip plus -1fil

\begin{IEEEbiography}[{\includegraphics[width=1in,height=1.25in,clip,keepaspectratio]{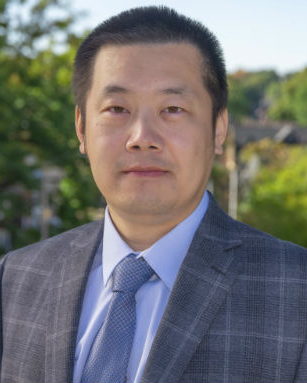}}]{Ziming Zhang} is an assistant professor and the P. I. of the Vision, Intelligence, and System Laboratory (VISLab) at Worcester Polytechnic Institute, 100 Institute Rd, Worcester, MA 01609, USA. Before joining WPI, he was a research scientist at Mitsubishi Electric Research Laboratories (MERL). Prior to that, he was a research assistant professor at Boston University. Dr. Zhang received his Ph.D. in 2013 from Oxford Brookes University, UK. (e-mail: zzhang15@wpi.edu).
\end{IEEEbiography}

\vskip -2\baselineskip plus -1fil

\begin{IEEEbiography}[{\includegraphics[width=1in,height=1.25in,clip,keepaspectratio]{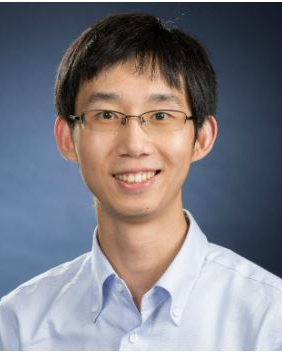}}]{Haichong K. Zhang} is an assistant professor and the P. I. of the Frontier Ultrasound Imaging and Robotic Instrumentation Laboratory (Medical FUSION Lab) at Worcester Polytechnic Institute, 100 Institute Rd, Worcester, MA 01609, USA. He holds a Ph.D. and an M. S. in Computer Science from Johns Hopkins University, Baltimore, MD 21218, USA. He has an M.S. and a B. S. in Health Sciences from Kyoto University, Kyoto, Japan. (e-mail: hzhang10@wpi.edu).
\end{IEEEbiography}
\vfill

\end{document}